\documentclass[12pt,letterpaper]{article}
\usepackage[a4paper, total={7in, 10in}]{geometry}

\usepackage{graphicx}
\usepackage{helvet}
\usepackage{authblk}
\usepackage{hyperref}
\usepackage{amsmath} 
\usepackage{amssymb} 
\usepackage{orcidlink} 
\usepackage[super,comma,sort&compress]  
   {natbib}
\usepackage[title]{appendix}%

\usepackage[utf8]{inputenc} 
\usepackage[T1]{fontenc}    
\usepackage{url}            
\usepackage{booktabs}       
\usepackage{amsfonts}       
\usepackage{nicefrac}       
\usepackage{microtype}      
\usepackage{xcolor}         
\usepackage{multirow}
\usepackage{threeparttable}
\usepackage{amsmath}
\usepackage{subcaption}
\usepackage{makecell}
\usepackage{placeins}
\usepackage[utf8]{inputenc} 
\usepackage{booktabs}      
\usepackage{siunitx}       
\usepackage{caption}       
\usepackage{tabularx}
\usepackage{listings}    
\usepackage{pifont}
\usepackage{array}
\usepackage{makecell}
\usepackage{float} 
\theadset{\bfseries}

\makeatletter
\renewcommand{\maketitle}{\bgroup\setlength{\parindent}{0pt}
\begin{flushleft}
  \textbf{\@title}
  
  \@author
\end{flushleft}\egroup}
\makeatother

\title{ECG2Gene: Deep Learning-Based Germline Tumor Susceptibility Gene Risk Assessment Using Electrocardiograms}
\date{}

\author[1,4,\# \orcidlink{0009-0000-1986-8131}]{Shanwei Zhang}
\author[2,\#]{Qinghao Zhao}
\author[3,\#]{Yize Zhao}
\author[4,5]{Jun Li}
\author[1,4,\orcidlink{0009-0003-8345-2212}]{Yongchao Long}
\author[1,4]{Jinshuai Gu} 
\author[6]{Ling Yang}
\author[6]{Huaidong Du}
\author[6]{Yiping Chen}
\author[6]{Zhengming Chen}
\author[3,7,8]{Dianjianyi Sun}
\author[3,7,8,9]{Jun Lv}
\author[3,7,8]{Liming Li}
\author[10]{Iona Y. Millwood}
\author[10]{Robin G.Walters}
\author[3,7,8,*]{Canqing Yu}
\author[1,11,*]{Yuxi Zhou}
\author[1,*]{Shenyong Chen}
\author[4,5,*,\orcidlink{0000-0001-7521-5127}]{Shenda Hong}

\affil[1]{Department of Computer Science, Tianjin University of Technology, Tianjin, China}
\affil[2]{Department of Cardiology, Peking University People’s Hospital, Beijing, China}
\affil[3]{Department of Epidemiology \& Biostatistics, School of Public Health,
Peking University, Beijing 100191, China}
\affil[4]{National Institute of Health Data Science, Peking University, Beijing, China}
\affil[5]{Institute for Artificial Intelligence, Peking University, Beijing, China}
\affil[6]{Clinical Trial Service Unit \& Epidemiological Studies Unit (CTSU), Nuffield Department of Population Health, University of Oxford, United Kingdom}
\affil[7]{Peking University Center for Public Health and Epidemic Preparedness \& Response, Beijing 100191, China;}
\affil[8]{Key Laboratory of Epidemiology of Major Diseases (Peking University), Ministry of Education, Beijing 100191, China}
\affil[9]{State Key Laboratory of Vascular Homeostasis and Remodeling, Peking University, Beijing 100191, China}
\affil[10]{Clinical Trial Service Unit \& Epidemiological Studies Unit (CTSU), Nuffield Department of Population Health, University of Oxford, United Kingdom}
\affil[11]{DCST, BNRist, RIIT, Institute of Internet Industry, Tsinghua University, Beijing, China}

\affil[*]{Correspondence: yucanqing@pku.edu.cn, joy\_yuxi@pku.edu.cn, csy@tjut.edu.cn, hongshenda@pku.edu.cn}

\begin{document}

\maketitle

\section*{ABSTRACT}



Electrocardiograms (ECGs) may contain physiological information related to hereditary cancer susceptibility, but its relationship with future cancer risk remains unclear. Here, we present ECG2Gene, a two-stage deep-learning framework that derives digital phenotypes related to germline tumor susceptibility from routine 12-lead ECGs and predicts future cancer risk. In Stage 1, carrier status for pathogenic or likely pathogenic variants in ATM, BRIP1, CHEK2, PALB2, BRCA1, and BRCA2 supervises a multitask model that generates six continuous ECG-derived gene scores. In Stage 2, XGBoost integrates these scores with clinical variables to predict subsequent breast and ovarian cancer in women and prostate cancer in men. We also introduce ECG-Wide Association Analysis (EWAS) to characterize associations between ECG-derived gene scores and ECG waveforms. Across the clinical feature combinations evaluated in UK Biobank, the best-performing sex-specific models achieve AUROCs of 0.683, 0.633, and 0.660 for breast, ovarian, and prostate cancer, respectively. In external evaluations, models combining ECG-derived gene scores with age achieve prostate cancer AUROCs of 0.709 and 0.579 among men in the overall China Kadoorie Biobank and MIMIC-IV-ECG cohorts, respectively. ECG2Gene links genetically supervised ECG phenotyping and waveform association analysis to future cancer outcomes, demonstrating the potential of routine ECGs combined with clinical information for non-invasive, opportunistic cancer risk assessment.
\section*{KEYWORDS}

ECG; Gene; Deep learning; Digital phenotyping; Cancer risk prediction


\section*{INTRODUCTION}

Hereditary cancer susceptibility carriers need to be identified before cancer develops. Pathogenic or likely pathogenic germline variants in BRCA1, BRCA2 increase risk for breast, ovarian, prostate, and other cancers\cite{kuchenbaecker2017brca12, xiong2025breast}. Identifying these carriers guides genetic counselling, intensified screening, risk-reducing interventions, and cascade testing in families. Yet population-level identification remains incomplete. Cost, access, uptake, family-history ascertainment, and variants of uncertain significance limit genetic testing in routine screening settings. Functional studies of BRCA2 further show that clinical variant classification requires rigorous annotation and biological evidence in addition to variant detection\cite{huang2025functionalNature}. These limitations leave a practical gap between individuals who could benefit from genetic risk assessment and the people who actually receive it early enough.


Electrocardiography offers a scalable phenotyping layer for this gap. A standard 12-lead ECG is inexpensive, non-invasive, widely available, and often collected before cancer diagnosis during routine care, preoperative assessment, hospitalization, and health screening. AI-ECG studies have shown that ECGs encode more than rhythm and interval measurements: deep learning can detect left ventricular dysfunction, atrial fibrillation risk during sinus rhythm, age, sex, and broad disease-related representations from ECG signals\cite{attia2019screening,attia2019AF,attia2019age,friedman2025unsupervised}. Large-scale ECG genetics and genotype-to-ECG studies also show that ECG morphology has a genetic basis and can serve as a measurable intermediate phenotype between inherited background and disease risk \cite{verweij2020geneticcell,lin2026empowering_capecg}. These findings support ECG as a testable source of genetic and systemic physiological signals, even when the target disease originates outside the heart.


We frame the biological link between tumor susceptibility genes and ECG as an indirect phenotype-to-gene signal. Germline variants exist in all nucleated cells, including cardiomyocytes. Genes such as ATM, CHEK2, and PALB2 participate in DNA damage repair, oxidative stress response, cell-cycle regulation, and cell survival, which also influence myocardial homeostasis and stress response. Mechanistic and clinical studies suggest that BRCA1/2 biology can extend beyond cancer target tissues and relate to myocardial injury, cardiovascular risk, and treatment-associated cardiotoxicity \cite{shukla2011brca1,gast2018cardiovascular_brca12,sajjad2017brca12_mutation}. This evidence supports a hypothesis: germline tumor susceptibility can leave subtle, systemic, and age or sex dependent physiological traces that deep learning can extract from ECG.


Eye2Gene provides the closest methodological analogy for this study. In inherited retinal diseases, Pontikos et al. showed that multimodal retinal imaging can support genetic diagnosis by learning gene-specific phenotypic patterns, achieving externally validated top-five accuracy of 83.9\% across 63 genetic causes and improving phenotype-driven variant prioritization \cite{pontikos2025eye2gene}. Eye2Gene expands the ability to recognize genotype-linked phenotypes and prioritize molecular interpretation. ECG2Gene tests a related phenotype-to-gene idea in a harder setting: using a routine, non-cancer, non-genomic physiological signal to learn representations of germline tumor susceptibility genes and assess their value for future cancer risk stratification.


Existing work has established the pieces of this question, while hereditary cancer risk remains untested as a phenotype-to-gene ECG task. Cancer genetics studies define the clinical importance of germline susceptibility genes \cite{kuchenbaecker2017brca12,hu_population_breast_cancer,huang2025functionalNature}. ECG genetics and AI-ECG studies show that ECG can encode inherited and systemic physiological information \cite{attia2019screening,verweij2020geneticcell,attia2019AF,attia2019age,friedman2025unsupervised,zhang2026ecgomics,mckeen2025ecgfm,lin2026empowering_capecg}. Phenotype-to-gene studies demonstrate that clinical phenotypes can aid genetic diagnosis and variant prioritization \cite{pontikos2025eye2gene,radhakrishnan2023cross}. Risk models such as BOADICEA integrate family history, pathogenic variants, PRS, and questionnaire risk factors for cancer risk prediction \cite{petitjean2025baseline,yang2024validation}. The remaining question is whether a routine ECG can produce an ECG-derived gene score that captures part of the germline susceptibility-related phenotype and contributes to breast, ovarian, and prostate cancer risk stratification.


To overcome these challenges, we propose ECG2Gene, a two-stage framework for ECG-derived germline tumor susceptibility phenotyping.In the first stage, ECG2Gene uses 12-lead ECGs to learn gene scores for cancer susceptibility genes, including ATM, BRIP1, CHEK2, PALB2, BRCA1, and BRCA2. In the second stage, these ECG derived gene scores are combined with readily available clinical variables, including age, sex, BMI, and smoking status, to predict future risk of breast, ovarian, and prostate cancer. The main contributions of this study are as follows:

\begin{itemize}
    \item ECG2Gene, a two-stage framework that learns continuous scores for six hereditary cancer susceptibility genes from routine 12-lead ECGs and integrates them with readily available clinical variables. This framework establishes a low-cost, non-invasive, and scalable approach to opportunistic screening and risk stratification for hereditary cancer susceptibility.
    \item 
    External validation in two independent cohorts, MIMIC and CKB, shows that ECG-derived gene scores, combined with clinical information, provide modest predictive and risk-stratification value for selected cancer outcomes.
    \item Introducing EWAS (ECG-Wide Association Analysis) to explore potential associations between genes and ECG phenotypes. We introduce EWAS to systematically examine associations between ECG-derived gene scores and ECG waveforms. By integrating analyses of subsequent cancer outcomes, we explore potential links between gene-related ECG phenotypes and cancer risk.
\end{itemize}

\section*{RESULTS}
\subsection*{ECG-Derived Gene Scores Combined with Clinical Variables Predict Cancer Risk}

\begin{figure}[!ht] 
\centerline{\includegraphics[width=1\textwidth]{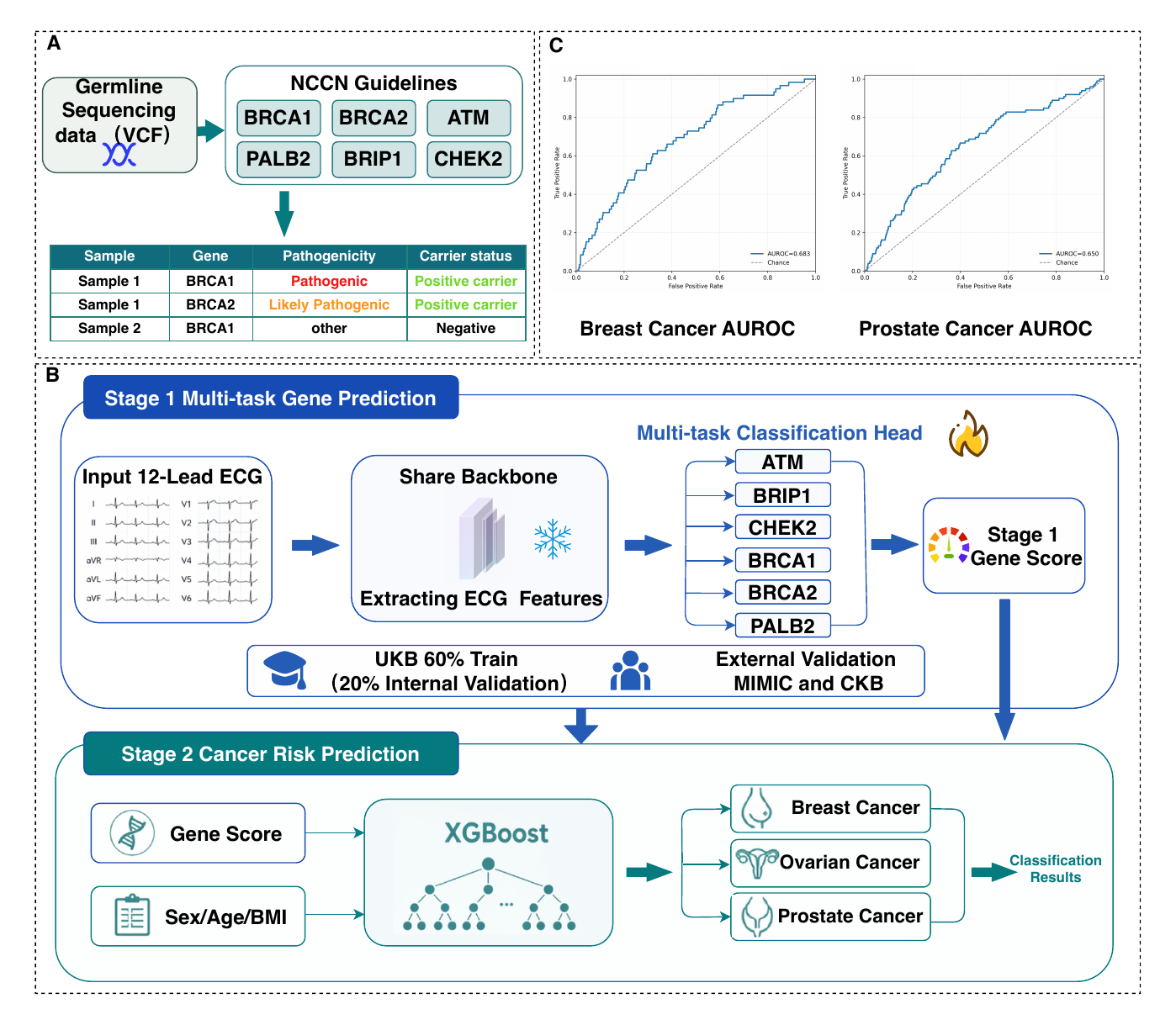}}

\caption{Framework Overview: A, Construction of Positive Gene Carrier Labels and Variant Pathogenicity Annotation. B, Two-Stage ECG2Gene Framework. C, AUROC for Breast Cancer in Female; Right: AUROC for Prostate Cancer in Male.
}
\label{ecg-gene-workflow}
\end{figure}

Sex-specific analyses in the UK Biobank showed that combining six ECG-derived gene scores for \textit{ATM}, \textit{BRCA1}, \textit{BRCA2}, \textit{BRIP1}, \textit{CHEK2}, and \textit{PALB2} with age and BMI in XGBoost yielded higher AUROCs for breast, ovarian, and prostate cancer than models using either the gene scores or clinical variables alone (Figure~\ref{ecg-gene-workflow}C; Table~\ref{tab:baseline_comparison_breast_ukb}). For breast cancer among female participants, the model combining the gene scores with age and BMI achieved an AUROC of 0.669 (95\% CI: 0.603--0.735), compared with 0.516 (95\% CI: 0.447--0.585) for the gene-score-only model and 0.641 (95\% CI: 0.566--0.716) for the clinical model comprising age, BMI, and smoking status. For prostate cancer among male participants, the corresponding AUROC was 0.660 (95\% CI: 0.603--0.717), compared with 0.581 (95\% CI: 0.526--0.636) for the ECG-drivied gene score model and 0.616 (95\% CI: 0.562--0.670) for the clinical model. The left and right panels of Figure~\ref{ecg-gene-workflow}C present the ROC curves for predicting breast cancer among female participants and prostate cancer among male participants, using the six ECG-derived gene scores combined with age and BMI.

For ovarian cancer among female participants, combining the ECG-derived gene scores with age and BMI yielded an AUROC of 0.604 (95\% CI: 0.356--0.852). Combining the scores with age and smoking status produced the highest AUROC of 0.633 (95\% CI: 0.392--0.874), although the confidence interval was wide. After smoking status is added to the models containing the ECG-derived gene scores, age, and BMI, the AUROC for breast cancer increased to 0.683 (95\% CI: 0.618--0.748), whereas the corresponding AUROCs for ovarian and prostate cancer are 0.580 (95\% CI: 0.316--0.844) and 0.650 (95\% CI: 0.594--0.706), respectively. The full feature set do not yield the highest AUROC for every cancer outcome.

Results among White participants show a numerical pattern consistent with that in the overall population (Table~\ref{tab:ukb_auc_white}). Models combining the ECG-derived gene scores with age and BMI achieve AUROC of 0.666 (95\% CI: 0.598--0.734), 0.605 (95\% CI: 0.347--0.863), and 0.668 (95\% CI: 0.612--0.724) for breast, ovarian, and prostate cancer. These estimates are close to the corresponding AUROCs of 0.669 (95\% CI: 0.603--0.735), 0.604 (95\% CI: 0.356--0.852), and 0.660 (95\% CI: 0.603--0.717) in the overall population. Adding smoking status yields AUROC of 0.680 (95\% CI: 0.614--0.746) for breast cancer and 0.657 (95\% CI: 0.601--0.713) for prostate cancer. For ovarian cancer, the model combining the ECG-derived gene scores with age and smoking status retains the highest AUROC of 0.633 (95\% CI: 0.384--0.882).

Among participants with CVD at baseline, model performance varies across cancer outcomes (Table~\ref{tab:ukb_auc_cvd}). For breast cancer among female participants, the ECG-derived gene-score-only model achieves an AUROC of 0.710 (95\% CI: 0.547--0.873). Adding age increases the AUROC to 0.749 (95\% CI: 0.612--0.886), while combining the scores with age and BMI yields an AUROC of 0.737 (95\% CI: 0.589--0.885). Combining the scores with age and smoking status yields an AUROC of 0.749 (95\% CI: 0.619--0.879). For prostate cancer among male participants, the model combining the ECG-derived gene scores with age and BMI achieves the highest AUROC of 0.607 (95\% CI: 0.525--0.689). This estimate is close to the AUROC of 0.601 (95\% CI: 0.519--0.683) for the clinical model comprising age and BMI and exceeds the AUROC of 0.527 (95\% CI: 0.439--0.615) for the ECG-derived gene-score-only model. This subgroup contains no ovarian cancer events, precluding estimation of the AUROC for ovarian cancer.

Among participants with CVD at baseline, model performance varies across cancer outcomes (Table~\ref{tab:ukb_auc_cvd}). For breast cancer among female participants, the ECG-derived gene-score-only model achieves an AUROC of 0.710 (95\% CI: 0.547--0.873). Adding age increases the AUROC to 0.749 (95\% CI: 0.612--0.886), while combining the scores with age and BMI yields an AUROC of 0.737 (95\% CI: 0.589--0.885). Combining the scores with age and smoking status yields an AUROC of 0.749 (95\% CI: 0.619--0.879). For prostate cancer among male participants, the model combining the ECG-derived gene scores with age and BMI achieves the highest AUROC of 0.607 (95\% CI: 0.525--0.689). This estimate is close to the AUROC of 0.601 (95\% CI: 0.519--0.683) for the clinical model comprising age and BMI and exceeds the AUROC of 0.527 (95\% CI: 0.439--0.615) for the ECG-derived gene-score-only model. This subgroup contains no ovarian cancer events, precluding estimation of the AUROC for ovarian cancer.


Age provided the most consistent incremental discriminatory information in both the overall population and White participants, while BMI further increased the AUROC for breast and prostate cancer. The additional contribution of smoking status varied across cancer outcomes and feature combinations and is most evident for ovarian cancer.

\subsection*{ECG-Derived Gene Scores and Pathogenic Variants Inputs Show Comparable Discrimination}
In the UK Biobank breast cancer analysis, the BOADICEA model, which integrates family history, polygenic risk scores (PRS), questionnaire-based risk factors, and Pathogenic Variants (PV), achieves an AUROC of 0.660 (95\% CI: 0.660--0.670) among White women~\cite{petitjean2025baseline}. The model combining the ECG-derived gene scores with age, BMI, and smoking status achieves AUROCs of 0.683 (95\% CI: 0.618--0.748) among all women and 0.680 (95\% CI: 0.614--0.746) among White women (Table~\ref{tab:baseline_comparison_breast_ukb}), which are numerically similar to the BOADICEA estimate. Results are also consistent between the overall population and the White subgroup. Models directly incorporating PVs with age, BMI, and smoking status yield breast cancer AUROCs of 0.648 (95\% CI: 0.574--0.722) and 0.646 (95\% CI: 0.571--0.721), respectively. The corresponding models combining the ECG-derived gene scores with the same clinical variables yield AUROCs of 0.683 (95\% CI: 0.618--0.748) and 0.680 (95\% CI: 0.614--0.746), respectively (Tables~\ref{tab:ukb_pv_auc} and~\ref{tab:ukb_auc_white_pv}).


Among White women, most participants have a negative PV carrier status, which may limit the discriminatory information available to the PV-only model. The ECG-derived gene-score-only model achieves an AUROC of 0.509 (95\% CI: 0.438--0.580). With clinical variables in the models, the AUROC estimates for the two input types are within a similar range. Prostate cancer shows a similar numerical pattern in models incorporating clinical variables. Models combining Pathogenic Variants or ECG-derived gene scores with clinical variables achieve AUROCs of 0.625 (95\% CI: 0.572--0.678) and 0.650 (95\% CI: 0.594--0.706), among all men. The corresponding AUROCs among White men are 0.627 (95\% CI: 0.574--0.680) and 0.657 (95\% CI: 0.601--0.713), respectively (Tables~\ref{tab:ukb_pv_auc} and~\ref{tab:ukb_auc_white_pv}). For ovarian cancer, the full model using direct Pathogenic Variant inputs achieves a higher AUROC than the full model using ECG-derived gene scores, although both estimates have wide confidence intervals. Restricting the analysis to White participants changes the AUROCs by no more than 0.007 for the full ECG-derived gene-score models and by no more than 0.002 for the direct Pathogenic Variant models across the three cancer outcomes. Thus, restricting the analysis to White participants does not materially alter the AUROC point estimates or the relative ranking of the models.

\begin{table}[htbp]
\centering
\caption{Comparison with baseline breast cancer risk prediction models in UK Biobank}
\label{tab:baseline_comparison_breast_ukb}
\setlength{\tabcolsep}{5pt}
\renewcommand{\arraystretch}{1.25}

\begin{threeparttable}
\begin{tabularx}{\textwidth}{
|>{\centering\arraybackslash}m{0.24\textwidth}
|>{\centering\arraybackslash}X
|>{\centering\arraybackslash}m{0.22\textwidth}|
}
\hline
\textbf{Baseline} &
\textbf{Population / Model input} &
\makecell{\textbf{Breast cancer}\\\textbf{AUROC}\\\textbf{($\pm$ 95\% CI)}} \\
\hline

\multirow{5}{=}{\centering\makecell{BOADICEA\\FH + PRS + QRF + PV}} 
& White women 
& $0.66 \pm 0.005$ \\
\cline{2-3}

& White women aged younger than 50 years 
& $0.66 \pm 0.015$ \\
\cline{2-3}

& White women aged 50 years and older 
& $0.66 \pm 0.010$ \\
\cline{2-3}

& White pathogenic variant carriers 
& $0.66 \pm 0.025$ \\
\cline{2-3}

& Non-White women 
& $0.54 \pm 0.035$ \\
\hline

\multirow{8}{=}{\centering ECG2Gene}
& Women (Age + BMI + Smoke) 
& $0.641 \pm 0.075$ \\
\cline{2-3}

& Women (ECG-derived Gene Score) 
& $0.516 \pm 0.069$ \\
\cline{2-3}

& \makecell{Women (ECG-derived Gene Score\\+ Age + BMI + Smoke)}
& $0.683 \pm 0.065$ \\
\cline{2-3}

& White women (Age + BMI + Smoke)
& $0.637 \pm 0.076$ \\
\cline{2-3}

& White women (ECG-derived Gene Score)
& $0.509 \pm 0.071$ \\
\cline{2-3}

& \makecell{White women (ECG-derived Gene Score\\+ Age + BMI + Smoke)}
& $0.680 \pm 0.066$ \\
\hline

\multirow{2}{=}{\centering PV + Clinical}
& Women (PV + Age + BMI + Smoke) 
& $0.648 \pm 0.074$ \\
\cline{2-3}

& White women (PV + Age + BMI + Smoke) 
& $0.646 \pm 0.075$ \\
\hline

\end{tabularx}

\begin{tablenotes}
\footnotesize
\item FH = family history; QRF = questionnaire-based risk factors; PRS = polygenic risk scores; PV = pathogenic variants.
\end{tablenotes}

\end{threeparttable}
\end{table}

\subsection*{ECG2Gene Stratifies Cancer Risk in UK Biobank and China Kadoorie Biobank}

\begin{figure}[!ht]
\centering
\includegraphics[width=\textwidth]{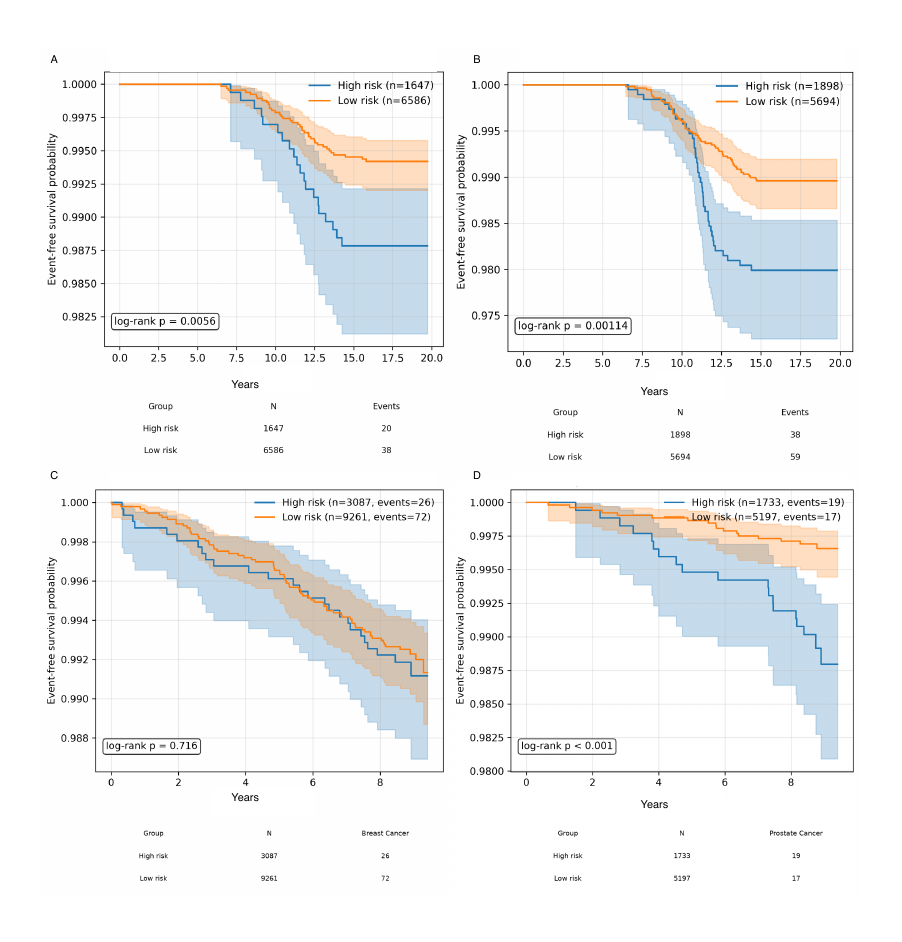}
\caption{
(A) Breast cancer among female UK Biobank participants
(B) Prostate cancer among male UK Biobank participants
(C) Breast cancer among female China Kadoorie Biobank participants
(D) Prostate cancer among male China Kadoorie Biobank participants
}
\label{figure-km-ukb-ckb}
\end{figure}


XGBoost uses the six ECG-derived gene scores, age, and BMI to generate outcome-specific risk scores that stratify participants into high- and low-risk groups for breast and prostate cancer in the UK Biobank and China Kadoorie Biobank. Figure~\ref{figure-km-ukb-ckb} presents the corresponding Kaplan--Meier curves. Among women in the UK Biobank, the high-risk group has a lower breast cancer-free survival probability than the low-risk group. Breast cancer occurs in 20 of 1,647 participants in the high-risk group and 38 of 6,586 participants in the low-risk group. Figure~\ref{figure-km-ukb-ckb}A shows separation between the Kaplan--Meier curves with a log-rank \(P=0.0056\). Among men in the UK Biobank, 38 prostate cancer events occur among 1,898 participants in the high-risk group, whereas 59 events occur among 5,694 participants in the low-risk group. Figure~\ref{figure-km-ukb-ckb}B shows a more rapid decline in prostate cancer-free survival probability in the high-risk group with a log-rank \(P=0.00114\).

The Kaplan--Meier curves in the China Kadoorie Biobank show cancer-specific risk-stratification patterns. In the breast cancer analysis among women, 26 events occur among 3,087 participants in the high-risk group, while 72 events occur among 9,261 participants in the low-risk group. Figure~\ref{figure-km-ukb-ckb}C shows substantial overlap between the two curves, and the log-rank test detects no difference in breast cancer-free survival (\(P=0.716\)). In the prostate cancer analysis among men, 32 events occur among 4,820 participants in the high-risk group, while four events occur among 14,458 participants in the low-risk group. Figure~\ref{figure-km-ukb-ckb}D shows clear separation between the Kaplan--Meier curves with a log-rank \(P<0.001\). These findings show that the ECG2Gene risk scores stratify prostate cancer risk in the China Kadoorie Biobank, whereas they do not significantly stratify breast cancer risk in this cohort.

The risk-stratification performance of individual ECG-derived gene scores varies by gene and cancer outcome. Among women in the UK Biobank, only the \textit{PALB2} score separates the high- and low-risk curves when breast or ovarian cancer serves as the event. Figure~\ref{km-ukb-female}E shows this separation with a log-rank \(P=0.0203\). The other five gene scores do not show clear curve separation. Among men in the UK Biobank, only the \textit{BRCA2} score differentiates the high- and low-risk groups for prostate cancer. Figure~\ref{km-ukb-male}B shows this difference with a log-rank \(P=0.00115\), whereas the remaining five scores show no clear risk-stratification patterns. Cancer risk scores that combine the six ECG-derived gene scores, age, and BMI show more consistent risk stratification than the individual gene scores. The individual-score risk-stratification signals occur primarily for the \textit{PALB2} score among women and the \textit{BRCA2} score among men.

\subsection*{External Validation of ECG2Gene}

\begin{table}[htbp]
\centering
\caption{AUROC Performance of ECG-Derived Gene Scores Across Cohorts and CVD Subgroups}
\label{tab:external_validation_cvd_subgroups}

\setlength{\tabcolsep}{5pt}
\renewcommand{\arraystretch}{1.15}
\footnotesize

\begin{threeparttable}
\begin{tabularx}{\textwidth}{
@{}
>{\raggedright\arraybackslash}p{0.18\textwidth}
>{\centering\arraybackslash}X
>{\centering\arraybackslash}X
>{\centering\arraybackslash}X
@{}
}
\toprule

\textbf{Cohort}
& \thead{\textbf{Overall}\\\textbf{AUROC}}
& \thead{\textbf{CVD}\\\textbf{AUROC}}
& \thead{\textbf{No CVD}\\\textbf{AUROC}} \\

\midrule

\multicolumn{4}{@{}l}{
\textbf{Female: Breast Cancer}
\quad
\textit{(ECG-Derived Gene Score + Age + BMI)}
} \\

UKB
& $0.669 \pm 0.065$
& $0.737 \pm 0.148$
& $0.654 \pm 0.076$ \\


CKB
& $0.536 \pm 0.058$
& $0.426 \pm 0.187$
& $0.558 \pm 0.053$ \\

\addlinespace[4pt]
\midrule

\multicolumn{4}{@{}l}{
\textbf{Male: Prostate Cancer}
\quad
\textit{(ECG-Derived Gene Score + Age + BMI)}
} \\

UKB
& $0.660 \pm 0.058$
& $0.607 \pm 0.082$
& $0.668 \pm 0.076$ \\

MIMIC
& $0.558 \pm 0.009$
& $0.536 \pm 0.016$
& $0.618 \pm 0.0136$ \\

CKB
& $0.709 \pm 0.073$
& $0.805 \pm 0.113$
& $0.696 \pm 0.078$ \\

\bottomrule
\end{tabularx}

\begin{tablenotes}[flushleft]
\footnotesize
\item Values are reported as AUROC $\pm$ the half-width of the 95\% confidence interval. CVD denotes cardiovascular disease.
\end{tablenotes}

\end{threeparttable}
\end{table}

Further race-stratified analyses show that the breast cancer model has limited discriminatory performance among both White and non-White participants in MIMIC. Table~\ref{tab:mimic_breast_auc_race} presents these results. For ovarian cancer, combining the ECG-derived gene scores with age and BMI yields AUROC of 0.528, 0.543, and 0.557 among all women, White women, and non-White women, respectively, with a slightly higher point estimate among non-White women. Table~\ref{tab:mimic_ovarian_auc_race} reports the corresponding subgroup results. For prostate cancer, the AUROCs are 0.558, 0.545, and 0.589, respectively, with the highest point estimate among non-White men. Table~\ref{tab:mimic_prostate_auc_race} presents these estimates. These results indicate that racial subgroup differences in MIMIC comprise modest changes in AUROC point estimates and do not alter the general pattern of weaker performance for breast cancer and relatively higher performance for ovarian and prostate cancer.

CVD-stratified analyses in MIMIC show that discriminatory performance for prostate cancer remains primarily among participants without CVD. Combining the ECG-derived gene scores with age and BMI yields AUROCs of 0.465 (95\% CI: 0.450--0.480), 0.480 (95\% CI: 0.450--0.510), and 0.618 (95\% CI: 0.604--0.632) for breast, ovarian, and prostate cancer, respectively, in the subgroup without CVD. Table~\ref{tab:minic_auc_without_cvd} presents these results. In the subgroup with CVD, the corresponding AUROCs are 0.476 (95\% CI: 0.462--0.490), 0.527 (95\% CI: 0.489--0.565), and 0.536 (95\% CI: 0.525--0.547), respectively. Table~\ref{tab:minic_auc_with_cvd} presents the results for this subgroup. The models show limited discriminatory performance for breast and ovarian cancer among women in both CVD subgroups, whereas the AUROC for prostate cancer is higher among participants without CVD than among those with CVD.


In the overall CKB population, combining the ECG-derived gene scores with age and BMI yields AUROC of 0.536, 0.431, and 0.709 for breast, ovarian, and prostate cancer, respectively, with the strongest discrimination for prostate cancer. Table~\ref{tab:ckb_auc} presents these results. In the subgroup without CVD, the corresponding AUROCs are 0.558 (95\% CI: 0.505--0.611), 0.422 (95\% CI: 0.292--0.552), and 0.696 (95\% CI: 0.618--0.774), respectively. The prostate cancer estimate is consistent with the corresponding estimate in the overall CKB population. Table~\ref{tab:ckb_auc_without_cvd} presents the results for participants without CVD. In the subgroup with CVD, the AUROCs are 0.426 (95\% CI: 0.239--0.631) for breast cancer and 0.805 (95\% CI: 0.692--0.918) for prostate cancer. The ovarian cancer AUROC cannot be estimated because this subgroup contains no outcome events. The prostate cancer model retains a relatively high AUROC point estimate, although the confidence interval is wide. Table~\ref{tab:ckb_auc_with_cvd} presents the results for participants with CVD. Across the supplementary analyses, the external discriminatory performance of ECG2Gene is most evident for prostate cancer, particularly in the overall CKB population and the subgroup without CVD. External validation performance is limited for breast and ovarian cancer, and the association between CVD status and model performance shows no consistent direction.

\subsection*{Feature Importance of ECG-Derived Gene Scores}

\begin{figure}[!ht] %
\centerline{\includegraphics[width=1\textwidth]{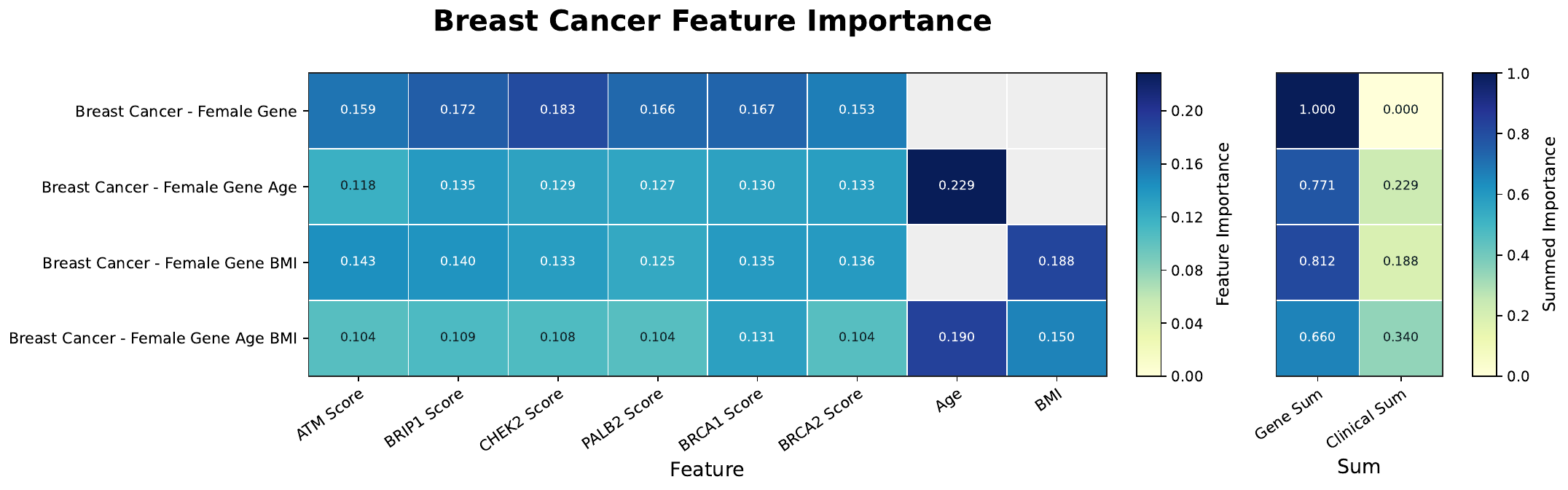}}
\caption{Feature importance in the female breast and ovarian cancer XGBoost risk prediction models. The left panel shows the importance of individual gene scores and clinical variables, and the right panel summarizes the overall importance of Gene Score and clinical variables.}
\label{feature_importance_female}
\end{figure}

\begin{figure}[!ht] %
\centerline{\includegraphics[width=1\textwidth]{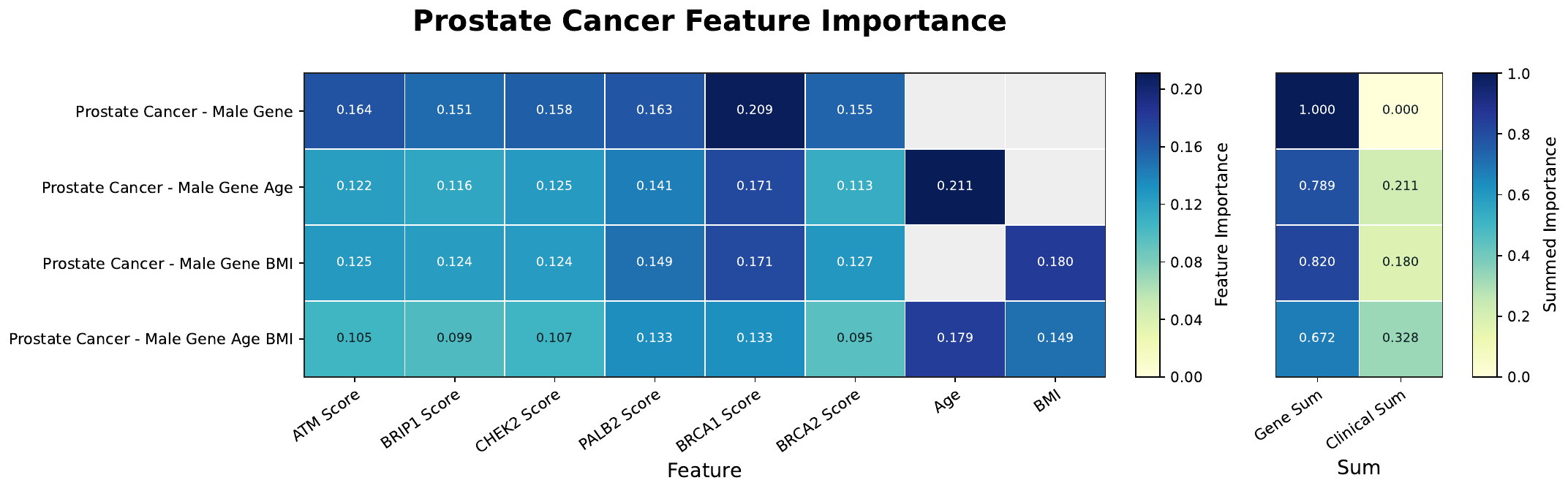}}
\caption{Feature importance in the male Prostate cancer XGBoost risk prediction models. The left panel shows the importance of individual gene scores and clinical variables, and the right panel summarizes the overall importance of Gene Score and clinical variables.}
\label{feature_importance_male}
\end{figure}

XGBoost feature-importance analysis shows that the six ECG-derived gene scores account for the majority of total feature importance in both the female breast cancer model and the male prostate cancer model. Figures~\ref{feature_importance_female} and~\ref{feature_importance_male} present the corresponding analyses for the female and male models, respectively. With only the gene scores in the models, individual feature-importance values range from 15.3\% to 18.3\% in the breast cancer model, and the \textit{CHEK2} score has the highest value. In the prostate cancer model, the corresponding values range from 15.1\% to 20.9\%, and the \textit{BRCA1} score has the highest value. Adding age results in combined feature-importance values of 77.1\% and 78.9\% for the six gene scores in the breast and prostate cancer models, respectively, while age accounts for 22.9\% and 21.1\%. Adding BMI results in combined feature-importance values of 81.2\% and 82.0\% for the gene scores, while BMI accounts for 18.8\% and 18.0\%, respectively. With both age and BMI in the breast cancer model, the six gene scores collectively account for 66.0\% of total feature importance, while age and BMI account for 19.0\% and 15.0\%, respectively. In the prostate cancer model, the corresponding values are 67.2\% for the six gene scores, 17.9\% for age, and 14.9\% for BMI. Thus, age has greater individual feature importance than BMI in both models, while the six ECG-derived gene scores collectively account for the majority of total feature importance across all feature combinations.



\subsection*{Associations Between ECG-Derived Gene Scores and 12-Lead ECG Waveforms}
\begin{figure}[!ht] 
\centerline{\includegraphics[width=1\textwidth]{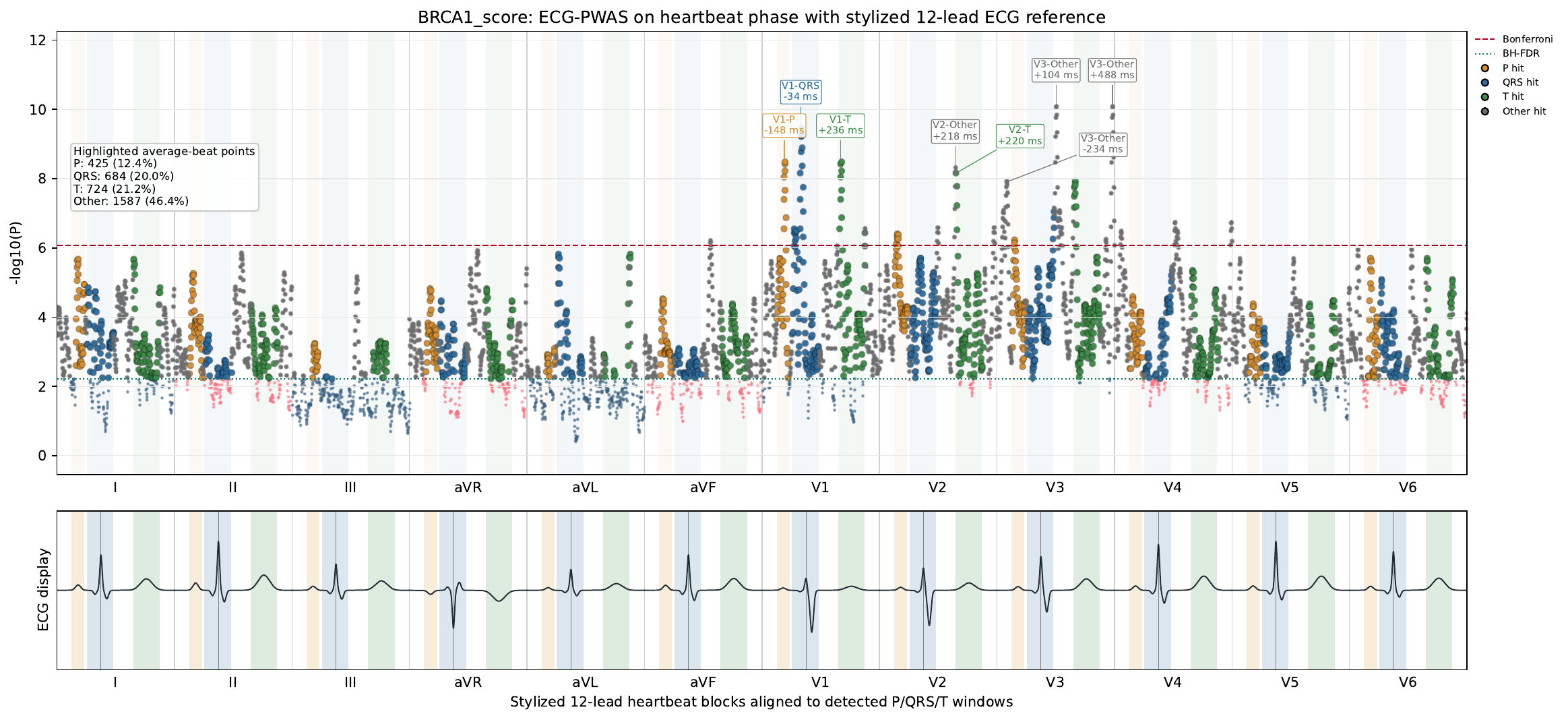}}
\caption{BRCA1 gene score-associated ECG time points across 12 leads.
}
\label{pwas_BRCA1}
\end{figure}

ECG-wide time-point association analysis identifies gene-score-specific waveform associations across 60,000 lead--time-point features from the 12 ECG leads. The numbers of associated time points meeting the FDR threshold (\(q<0.05\)) are 3,420, 3,106, 2,316, 1,483, 3,810, and 673 for the \textit{BRCA1}, \textit{ATM}, \textit{BRCA2}, \textit{CHEK2}, \textit{BRIP1}, and \textit{PALB2} scores, respectively. For all six scores, the largest proportion of associated time points occurs outside the P-wave, QRS-complex, and T-wave boundaries, accounting for 46.4\%, 45.0\%, 45.7\%, 50.6\%, 44.7\%, and 41.2\%, respectively. Among the three principal ECG waveform components, the T wave contains the largest proportion of associated time points for the \textit{BRCA1}, \textit{ATM}, \textit{BRCA2}, \textit{CHEK2}, and \textit{BRIP1} scores, accounting for 21.2\%, 23.3\%, 19.5\%, 17.9\%, and 24.5\%, respectively. In contrast, the QRS complex contains the largest proportion of associated time points for the \textit{PALB2} score, accounting for 24.1\%.


The \textit{BRCA1} score shows a broad association pattern across the precordial leads. Among the 3,420 time points that meet the FDR threshold, 425, 684, 724, and 1,587 occur within the P wave, QRS complex, T wave, and other cardiac phases, respectively. Of these time points, 167 also exceed the Bonferroni threshold (\(P<8.33\times10^{-7}\)). The associated time points cluster in leads V1--V4, with 395, 379, 337, and 330 FDR-associated time points in leads V3, V2, V1, and V4, respectively. The strongest associations occur in lead V3 outside the P-wave, QRS-complex, and T-wave boundaries, including regions approximately 104 and 488~ms after the R peak. A prominent peak also occurs within the QRS complex of lead V1 approximately 34~ms before the R peak, with additional association clusters around the P and T waves in lead V1. Figure~\ref{pwas_BRCA1} presents this association pattern.


The association signals for the \textit{ATM} score concentrate primarily in the precordial leads. The strongest peak occurs within the QRS complex of lead V2 approximately 66~ms after the R peak, with additional association clusters around the QRS complex and T wave in lead V3. Figure~\ref{pwas_ATM} presents the \textit{ATM} association pattern.
The \textit{BRCA2} score shows a localized association pattern around lead V2. Its strongest peak occurs outside the waveform boundaries approximately 248~ms before the R peak. Additional peaks that exceed the Bonferroni threshold occur within the P wave approximately 168~ms before the R peak and within the QRS complex approximately 68~ms after the R peak in lead V2. Figure~\ref{pwas_BRCA2} presents the \textit{BRCA2} association pattern.
The strongest association for the \textit{CHEK2} score occurs within the P-wave region of lead V1 approximately 190~ms before the R peak. Additional localized peaks occur within the T-wave region of lead V3 approximately 292 and 388~ms after the R peak and in the adjacent cardiac phases. Figure~\ref{pwas_CHEK2} presents the \textit{CHEK2} association pattern.
The \textit{BRIP1} score contains the largest number of FDR-associated time points. Its most prominent peaks occur within the QRS complexes of leads V2 and V3 approximately 76 and 86~ms after the R peak, respectively, with dense association clusters extending into the adjacent post-QRS phases. Figure~\ref{pwas_BRIP1} presents the \textit{BRIP1} association pattern.
The \textit{PALB2} score has relatively fewer associated time points, which cluster within the QRS complexes of leads V1, V5, and V6 and in other cardiac phases of leads V2--V3. These signals meet the FDR threshold but do not exceed the Bonferroni threshold. Figure~\ref{pwas_PALB2} presents the \textit{PALB2} association pattern.

The six ECG-derived gene scores show distinct lead--time-point association patterns. The strongest local associations are concentrated within the QRS complex, T wave, and adjacent cardiac phases in the precordial leads V1--V3, indicating that the waveform information capture by these scores varied across genes, leads, and cardiac phases.

\section*{DISCUSSION}
ECG2Gene provides a proof-of-concept two-stage framework. First, it derives six scores from prediagnostic 12-lead ECGs, using carrier status for six germline cancer susceptibility genes as supervision. It then evaluates the value of these scores for cancer risk prediction. In the UK Biobank, adding the scores to baseline covariates produces only a limited improvement in breast cancer discrimination over the corresponding clinical model. Kaplan--Meier analyses further separate high-risk and low-risk groups for breast and prostate cancer in both the UK Biobank and CKB. Performance varies across external cohorts: the models for female cancers perform close to chance in MIMIC and CKB, whereas the prostate cancer model retains a comparatively higher AUROC. The use of prediagnostic ECGs reduces the potential influence of cancer diagnosis and treatment. However, occult disease and residual confounding remain possible. The current evidence therefore supports risk enrichment rather than cancer diagnosis or inference of individual carrier status.

ECG2Gene makes three contributions to phenotype-guided cancer risk assessment. First, the two-stage framework extends “phenotype-to-gene” learning to routine physiological signals by deriving six continuous gene scores from 12-lead ECGs under germline carrier-status supervision and combining them with readily available clinical information. Second, we introduce EWAS to systematically characterize associations between ECG-derived gene scores and ECG waveforms. Together with analyses of subsequent cancer outcomes, EWAS provides an interpretable framework for exploring potential links between gene-related ECG phenotypes and cancer risk. Third, we evaluate the combined models in two independent external cohorts, covering breast and prostate cancer in CKB and prostate cancer in MIMIC. These results support modest predictive and risk-stratification value for ECG-derived gene scores combined with clinical information in selected cancer outcomes, while showing that performance varies across cancer types and cohorts.

The potential clinical and societal value of ECG2Gene lies in risk triage before formal genetic assessment. Sequencing-based assessment and models such as BOADICEA rely on pathogenic-variant annotations, family history, polygenic risk scores, or questionnaire data, whereas ECG2Gene uses existing ECG records and basic clinical variables. Within routine ECG workflows, the scores identify individuals for prioritization for genetic counseling, confirmatory testing, or intensified screening without directly classifying them as variant carriers. With validation, this approach broadens access to risk assessment in settings where sequencing resources or reliable family histories are limited, while germline genetic testing and guideline-based assessment remain the standards for diagnosis and management. Heterogeneity across external cohorts also indicates that deployment requires local calibration, informed-consent procedures, genetic-privacy safeguards, equitable follow-up, and quantification of false-positive and false-negative burdens. ECG2Gene therefore complements, rather than replaces, germline genetic testing or established cancer-screening programs.

Despite the large source cohorts, relatively few participants carry pathogenic or likely pathogenic variants, which limits the model's ability to learn gene-related ECG features. Second, we use only P/LP variants as positive labels and exclude variants of uncertain significance and other variants with insufficient evidence. The model therefore depends on current knowledge of variant pathogenicity and cannot cover all relevant variants or genes; this study includes only six genes. Paired data comprising standard 12-lead ECGs and reliable genetic test results remain limited and restrict gene-specific training, independent external validation, and cross-population evaluation. ECG2Gene therefore remains a proof-of-concept framework and requires further validation using larger paired datasets.

Future work focuses on improving label quality and expanding the scope of ECG2Gene. As ClinVar continues to evolve, more variants of uncertain significance receive updated classifications, and pathogenic or likely pathogenic annotations become more accurate, providing more reliable training labels. We also explore additional cancers and susceptibility genes to understand how germline pathogenic variants manifest across different organs. ECG2Gene provides an initial cross-organ exploration: a genetic variant acts like a ``seed'' that produces different ``flowers'' in different organ environments, with its effects potentially appearing in both ECG-recorded physiological signals and tumor risks across organs. Larger datasets linking ECGs with genetic testing are needed to test this hypothesis.

\newpage
\section*{METHODS}


\subsection*{Related Works}
\subsubsection*{Hereditary cancer genetics and risk prediction}

Germline pathogenic variants in BRCA1, BRCA2, PALB2, ATM, CHEK2, and related genes form a clinically actionable substrate for cancer prevention \cite{kuchenbaecker2017brca12,hu_population_breast_cancer}. Population studies and prospective carrier cohorts establish high and moderate risk genes, while functional variant studies emphasize that carrier labels depend on stringent annotation and evolving evidence \cite{huang2025functionalNature}. BOADICEA and related models operationalize this knowledge for risk prediction by integrating family history, pathogenic variants, PRS, and questionnaire risk factors \cite{petitjean2025baseline,yang2024validation}. These models provide a genetic-risk reference and depend on information that many opportunistic screening settings lack. ECG2Gene occupies an earlier layer: it asks whether a routinely collected physiological phenotype can enrich risk before full genetic risk assessment.

\subsubsection*{ECG genetics and Al-ECG phenotyping}

ECG is a genetically influenced high-dimensional phenotype. GWAS and genotype-to-ECG studies show that ECG traits connect inherited background to cardiac physiology and disease risk \cite{verweij2020geneticcell,lin2026empowering_capecg}. AI-ECG models further show that deep representations can identify latent physiology, including ventricular dysfunction, atrial fibrillation risk, age, sex, and broad disease profiles \cite{attia2019screening,attia2019AF,attia2019age,friedman2025unsupervised}. ECGomics and ECG foundation models extend this view by treating ECG as a reusable digital biomarker \cite{zhang2026ecgomics,mckeen2025ecgfm}. ECG2Gene extends this to germline tumor susceptibility genes and cancer-specific downstream risk.

\subsubsection*{Signal phenotype to gene learning}


Eye2Gene, GestaltMatcher-like approaches, and cross-modal cardiovascular representation studies show that clinical phenotypes can inform genetic diagnosis, variant prioritization, and discovery of gene-phenotype neighborhoods \cite{pontikos2025eye2gene,radhakrishnan2023cross}. Their core insight is that high-resolution phenotypes can carry gene-specific patterns that experts recognize only partially. ECG2Gene adapts this idea to a lower-signal setting. Retinal imaging directly observes the affected tissue in inherited retinal diseases; ECG observes a systemic physiological projection of cancer susceptibility biology. This makes the ECG2Gene task harder and makes a biomarker framing essential.


ECG2Gene connects these three literatures by testing ECG-derived germline susceptibility phenotyping for future cancer risk. Compared with genetic risk models, ECG2Gene uses ECG and basic clinical variables as inputs. Compared with general AI-ECG disease prediction, it introduces gene-specific intermediate scores to make the cancer-risk signal biologically interrogable. Compared with Eye2Gene, it studies a lower-signal setting in which ECG provides an indirect systemic phenotype of cancer susceptibility biology. ECG2Gene evaluates whether ECG can provide a low-cost, weak, and externally testable signal for prioritizing cancer risk assessment and future genetic evaluation.

\subsection*{Study participants}

We use the UK Biobank cohort for model development and internal evaluation. We split participants into training, validation, and test sets using a 60\%/20\%/20\% ratio, ensuring that records from the same participant remain within a single subset. The final cohort includes 79,129 participants, with 47,478 in the training set, 15,826 in the validation set, and 15,825 in the test set. Genetic and cancer outcome labels are distributed proportionally across the three subsets, with the validation and test sets each containing approximately 20\% of positive cases for most labels.

We perform external validation in the CKB, MIMIC. During data preprocessing, we exclude samples with missing key variables and participants with a cancer diagnosis before ECG acquisition. Baseline demographic and clinical characteristics, including cancer event frequencies, are summarized in Table~\ref{tab:baseline_characteristics}. This restriction reduces the influence of prior cancer and cancer treatment on ECG phenotypes, as cancer therapy can cause myocardial injury and alter ECG features. We then identify genes associated with hereditary susceptibility to breast, ovarian, and prostate cancers according to the NCCN guidelines~\cite{cheng2026geneticNCCN}. Using the RAP platform, we annotate participants with matched genotype and ECG data, identify pathogenic variant carriers, and generate the corresponding gene-level labels.

\begin{figure*}[htbp] %
\centerline{\includegraphics[width=1\textwidth]{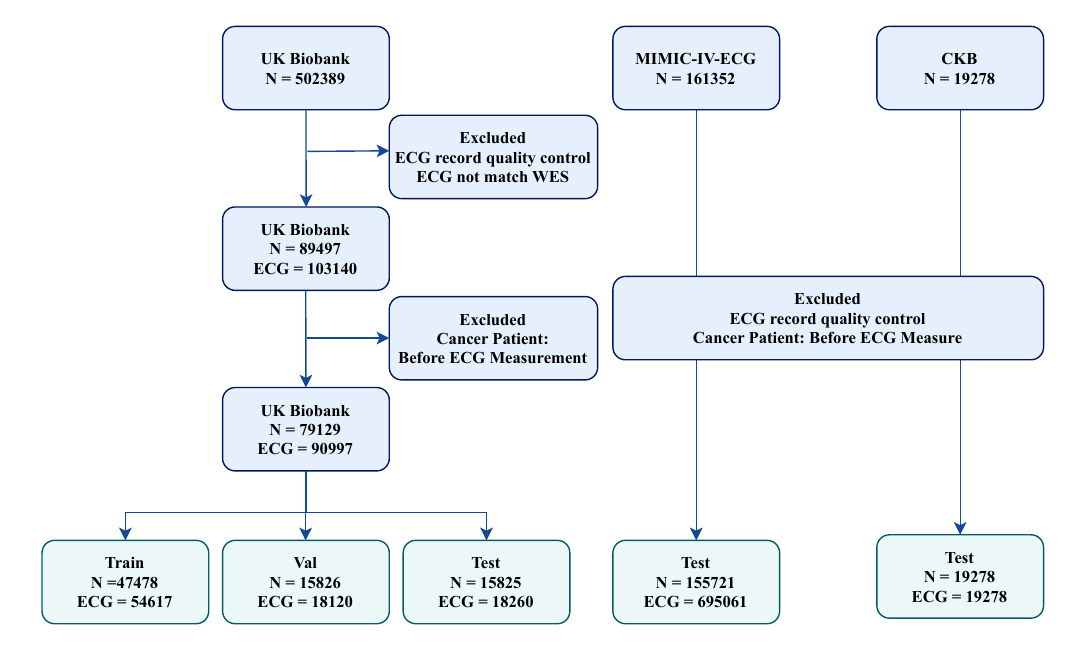}}
\caption{
Dataset 
}
\label{dataset}
\end{figure*}

\subsection*{Problem Definition} 
This study assesses whether routine ECG recorded before cancer diagnosis, together with a small set of readily available clinical variables, can predict an individual's future cancer risk. For participant \(i\), the ECG input is represented as
\begin{equation}
X_i \in \mathbb{R}^{12 \times 5000},
\end{equation}
where \(12\) denotes the standard 12 leads, and \(5000\) denotes the number of samples in a 10-second ECG recorded at 500 Hz. Because all ECGs were collected before cancer diagnosis, this task is defined as prospective risk prediction rather than cancer diagnosis or retrospective classification.

This study uses a two-stage modeling framework. The first stage learns ECG representations associated with pathogenic variant carrier status. We focus on six genes related to hereditary cancer risk:
\begin{equation}
G =
\left\{
\mathrm{ATM},
\mathrm{BRIP1},
\mathrm{CHEK2},
\mathrm{PALB2},
\mathrm{BRCA1},
\mathrm{BRCA2}
\right\}.
\end{equation}

For each gene \(g \in G\), we define a binary label:
\begin{equation}
y_{i,g}^{\mathrm{gene}} \in \{0,1\},
\end{equation}
where
\begin{equation}
y_{i,g}^{\mathrm{gene}} =
\begin{cases}
1, & \text{participant } i \text{ carries a pathogenic variant in gene } g, \\
0, & \text{participant } i \text{ is a non-carrier for gene } g.
\end{cases}
\end{equation}

We use a pretrained ECG-founder model as the ECG feature extractor. Under a transfer-learning strategy, we freeze the backbone parameters and train only gene-specific classification heads. For each gene \(g\), the model outputs an ECG-derived gene score:
\begin{equation}
s_{i,g} = f_g(X_i),
\end{equation}
where \(s_{i,g}\) denotes the ECG-derived score associated with pathogenic variant carrier status for gene \(g\). The first stage generates a six-dimensional gene score vector for each participant:
\begin{equation}
S_i =
\left[
s_{i,\mathrm{ATM}},
s_{i,\mathrm{BRIP1}},
s_{i,\mathrm{CHEK2}},
s_{i,\mathrm{PALB2}},
s_{i,\mathrm{BRCA1}},
s_{i,\mathrm{BRCA2}}
\right].
\end{equation}

In the second stage, we apply the trained first-stage models to all samples in the training, validation, and test sets to compute gene scores. We then combine these scores with clinical variables as inputs to the cancer risk prediction model. For participant \(i\), the second-stage feature vector is defined as
\begin{equation}
Z_i =
\left[
S_i,
\mathrm{age}_i,
\mathrm{sex}_i,
\mathrm{BMI}_i,
\mathrm{smoke}_i
\right],
\end{equation}
where \(\mathrm{age}_i\), \(\mathrm{sex}_i\), \(\mathrm{BMI}_i\), and \(\mathrm{smoke}_i\) denote age, sex, body mass index, and smoking status, respectively. Based on \(Z_i\), we train an XGBoost binary classifier to predict the incident cancer label:
\begin{equation}
y_i^{\mathrm{cancer}} \in \{0,1\},
\end{equation}
where
\begin{equation}
y_i^{\mathrm{cancer}} =
\begin{cases}
1, & \text{participant } i \text{ develops the target cancer during follow-up}, \\
0, & \text{participant } i \text{ does not develop the target cancer during follow-up}.
\end{cases}
\end{equation}

Because cancer risks differ by sex, we define cancer prediction tasks separately. Among female participants, the models predict breast cancer and ovarian cancer risk. Among male participants, the model predicts prostate cancer risk. We exclude male breast cancer from the analysis because UK Biobank contains no male breast cancer cases with both matched ECG data and positive pathogenic variant status.

The core question of this study can therefore be formalized as
\begin{equation}
P\left(y_i^{\mathrm{cancer}} = 1 \mid X_i, C_i\right),
\end{equation}
where
\begin{equation}
C_i =
\left[
\mathrm{age}_i,
\mathrm{sex}_i,
\mathrm{BMI}_i,
\mathrm{smoke}_i
\right].
\end{equation}

Specifically, this study evaluates whether ECG-derived gene scores,
\begin{equation}
S_i = F(X_i),
\end{equation}
combined with clinical features \(C_i\),
\begin{equation}
Z_i = \left[F(X_i), C_i\right],
\end{equation}
can predict an individual's future risk of breast cancer, ovarian cancer, or prostate cancer.

\subsection*{Dataset quality control and preparation}
\subsubsection*{Gene}

The study first screens candidate genes associated with hereditary susceptibility to breast cancer, ovarian cancer, and prostate cancer according to the NCCN Genetic/Familial High-Risk Assessment Guidelines. It then obtains the chromosomal coordinates of each candidate gene in the GRCh38 reference genome~\cite{grch38p14_ncbi} from the NCBI Gene database, including chromosome number, start position, and end position. ClinVar is further used to curate reported pathogenic or likely pathogenic variants in these genes.

Based on the genomic coordinates of the candidate genes, this study identifies the corresponding chromosome blocks in the UK Biobank whole-exome sequencing pVCF files and extracts variant data within the target gene regions for the study participants. The UK Biobank WES data use GRCh38 as the reference genome. To ensure coordinate and allele consistency between UK Biobank and ClinVar, all variants are normalized against GRCh38 and left-normalized, allowing direct comparison of the CHROM, POS, REF, and ALT fields.

Variant extraction is performed using bcftools. This study retains variants for which at least one participant in the study cohort carries a non-reference allele. Functional annotation is then performed on the UK Biobank RAP platform using SnpEff, with the annotation database based on the NCBI RefSeq GRCh38.p14 release. SnpSift Annotate is further used to integrate clinical significance annotations from the ClinVar GRCh38 VCF file into the variant file.

For quality control, this study standardizes, filters, and classifies the extracted variants. At the variant level, only variants meeting the predefined quality threshold, such as QUAL > 20, are retained, depending on the field definitions in the original VCF files. Variants annotated in ClinVar as Pathogenic, Likely pathogenic, or Pathogenic/Likely pathogenic are defined as pathogenic variants. Pathogenic variant classification is primarily based on exact matching of the left-normalized CHROM, POS, REF, and ALT fields, with additional strict verification using HGVS c. and HGVS p. annotations.

After candidate gene screening, we count the number of carriers of pathogenic or likely pathogenic variants in each gene. Because some genes have limited numbers of positive samples, which may increase class imbalance during ECG model training and limit the model's ability to learn stable gene-associated phenotypic features, we select six genes with sufficient carrier counts and established relevance to hereditary cancer susceptibility for subsequent analysis: ATM, BRCA1, BRCA2, BRIP1, CHEK2, and PALB2. Other genes are excluded from subsequent ECG-based gene prediction analyses because of insufficient numbers of pathogenic variant carriers. Each participant's gene status is defined according to whether the participant carries a ClinVar pathogenic or likely pathogenic variant in the target gene. If a participant carries at least one Pathogenic, Likely pathogenic, or Pathogenic/Likely pathogenic variant in a given target gene, that gene is labeled as positive; otherwise, it is labeled as negative. The number of positive samples for each gene is reported in Table~\ref{tab:ecg_positive_samples}.

To assess the reliability of pathogenic gene annotation in UK Biobank, we compare the pathogenic variant carrier frequencies obtained in this study with epidemiological reference frequencies reported in the CARRIERS population-based case-control study~\cite{hu_population_breast_cancer}. 
The CARRIERS study includes 32,247 breast cancer cases and 32,544 unaffected controls, and reports pathogenic variant carrier frequencies in cases and controls across multiple breast cancer predisposition genes. 
Overall, the BRCA1, BRCA2, ATM, and CHEK2 positive rates annotated in UK Biobank are within the same epidemiological range as prior population-based estimates. BRCA1, BRCA2, ATM, and CHEK2 carrier frequencies are higher in cases than in controls. As shown in Table~\ref{tab:gene_lp_p_distribution}, BRCA1 and BRCA2 show high breast cancer risk associations, with ORs of 11.71 (95\% CI: 6.47--21.18) and 5.61 (95\% CI: 3.87--8.14), respectively. ATM and CHEK2 show moderate risk associations, with ORs of 2.48 (95\% CI: 1.64--3.75) and 2.52 (95\% CI: 1.51--4.20), respectively. These results are consistent with the high-risk and moderate-risk gene stratification reported by Hu et al., indicating that the pathogenic gene annotation in UK Biobank recapitulates established epidemiological carrier frequencies and risk directions. This supports the reliability of our gene annotation pipeline and positive-label construction.

\subsubsection*{ECG Preprocess}
This study included only standard 12-lead ECG data with a sampling frequency of $500~\mathrm{Hz}$. Records with missing ECG file paths, unreadable files, empty waveform data, or incomplete lead information were directly excluded. The retained ECG data were organized according to the standard 12-lead, including leads I, II, III, aVR, aVL, aVF, and V1-V6.

Subsequently, each lead of each ECG record was preprocessed independently. First, a $1~\mathrm{Hz}$ high-pass filter was applied to remove baseline wander. Then, a $30~\mathrm{Hz}$ second-order Butterworth low-pass filter was used to suppress high-frequency noise. In addition, $50~\mathrm{Hz}$ and $60~\mathrm{Hz}$ notch filters were applied to attenuate power-line interference. All filtering procedures were performed using bidirectional zero-phase filtering to minimize phase distortion.

After filtering, each ECG record was uniformly processed to a duration of $10~\mathrm{s}$, corresponding to $5000$ sampling points. Finally, each lead of each ECG was standardized independently. For a single-lead signal $x$, its mean $\mu$ and standard deviation $\sigma$ were calculated, and the signal was transformed as follows:

\[x' = \frac{x - \mu}{d},\]

where
\[
d =
\begin{cases}
\sigma, & \sigma > 1 \times 10^{-8}, \\
1, & \sigma \leq 1 \times 10^{-8}.
\end{cases}
\]

This procedure was used to avoid numerical instability when the standard deviation was close to zero. Finally, each ECG record was represented as a $5000 \times 12$ matrix for subsequent model training and analysis.

\subsection*{Framework Overview}

\subsection*{Model Train}
\subsubsection*{Stage 1 : Predict ECG-Derived Gene Score}

Each ECG record was used as input to build a multi-task binary classification model for six gene labels: \textit{ATM}, \textit{BRIP1}, \textit{CHEK2}, \textit{PALB2}, \textit{BRCA1}, and \textit{BRCA2}. We split the dataset into training, validation, and test sets. 
The model used a pretrained Net1D as the feature-extraction backbone, with \(12\) input channels, convolutional filter configuration \([64, 160, 160, 400, 400, 1024, 1024]\), block configuration \([2, 2, 2, 3, 3, 4, 4]\), kernel size \(16\), and stride \(2\). 
We loaded weights from a \(1172\)-class pretraining task, froze the backbone during training, and optimized only the task-specific classification heads. 
Each classification head consisted of a \texttt{Linear-ReLU-Dropout-Linear} structure, with a hidden dimension of \(256\) and a dropout rate of \(0.3\), and produced the logit for the corresponding label. 
We train the model with binary Focal Loss using \(\alpha=0.9\) and \(\gamma_{\mathrm{FL}}=1.5\). Adam is used as the optimizer with a learning rate of \(1\times10^{-5}\), weight decay of \(1\times10^{-4}\), batch size of \(256\), and a maximum of \(30\) epochs. 
To reduce class imbalance, each training batch sampled any-label-positive samples and all-negative samples at a \(1:7\) ratio. Positive samples were augmented with a probability of \(0.9\) using amplitude scaling, random noise, temporal shifting, and temporal masking. 
The learning-rate scheduler monitored validation mean AUROC; when validation mean AUROC did not improve for \(2\) epochs, the learning rate was multiplied by \(0.5\), with a minimum learning rate of \(1\times10^{-6}\). 
Early stopping is triggered when validation mean AUROC failed to improve by at least \(1\times10^{-4}\) for \(8\) epochs. Model performance was evaluated using AUROC for each label, and the mean AUROC across the six labels was reported as the overall performance metric.

\subsubsection*{Stage 2 : Cancer Risk Prediction}
The current experimental configuration uses six gene prediction scores as input features, including \texttt{ATM\_score}, \texttt{BRIP1\_score}, \texttt{CHEK2\_score}, \texttt{PALB2\_score}, \texttt{BRCA1\_score}, and \texttt{BRCA2\_score}, without additionally including covariates such as age, BMI, smoking status, or sex. 
Under the female label preset, independent binary XGBoost models are trained for breast cancer and ovarian cancer. Under the male label preset, an independent binary model is trained for prostate cancer. 
Each model is trained as an independent binary classifier, with \texttt{auc} used as the evaluation metric during training. 
The main XGBoost hyperparameters are set as follows: the number of trees is \(300\), the learning rate is \(0.02\), the maximum tree depth is \(4\), \texttt{min\_child\_weight} is \(20\), \texttt{subsample} is \(0.5\), and \texttt{colsample\_bytree}, \texttt{colsample\_bylevel}, and \texttt{colsample\_bynode} are all \(0.5\). 
The \(L_1\) regularization coefficient is \(5\), the \(L_2\) regularization coefficient is \(20\), \texttt{gamma} is \(3\), \texttt{max\_delta\_step} is \(2\), \texttt{max\_bin} is \(64\), and \texttt{hist} is used as the tree construction method. 
To mitigate class imbalance, \texttt{scale\_pos\_weight} for each label is automatically calculated from the ratio between negative and positive samples in the training set, with the maximum value capped at \(10\). 
Model training uses the validation set for early stopping, and training stops if no improvement is observed for \(20\) consecutive rounds. 
Model performance is evaluated on the training, validation, and test sets. The primary metrics include AUROC and log loss for each label, and mean AUROC and average log loss across all labels are further reported. 
To assess result uncertainty, \(5000\) bootstrap resamples are used to calculate \(95\%\) confidence intervals.

\subsection*{Statistical Analysis}
Model performance is evaluated using metrics including AUROC and log loss. Uncertainty in model performance is estimated using nonparametric bootstrap resampling with 5,000 iterations. Performance metrics are recalculated in each bootstrap sample, and 95\% confidence intervals are derived from the 2.5th and 97.5th percentiles of the bootstrap distribution.


Because White participants provide most of the Stage 1 training data, we evaluate model performance separately in White and non-White groups to assess cross-population generalizability. We exclude participants with missing race or ethnicity from the race-stratified analysis. CKB serves as an East Asian external validation cohort without White versus non-White stratification. We also divide participants into CVD and no-CVD groups according to CVD status at ECG acquisition. In UKB and MIMIC, we classify participants as having CVD when any circulatory system disorder listed in UK Biobank Category 2409 occurs on or before the ECG date. In CKB, we define CVD as baseline acute myocardial infarction, angina, other ischemic heart disease, stroke, or transient ischemic attack. We apply the same trained models to all subgroups without retraining and report AUROC with 95\% confidence intervals.

Kaplan-Meier analysis is used to evaluate whether ECG-derived gene scores stratify subsequent cancer risk. For each of the six gene-specific scores, including ATM, BRIP1, CHEK2, PALB2, BRCA1, and BRCA2, participants in the highest quartile are defined as the high-risk group, and all remaining participants are defined as the low-risk group. Follow-up starts on the date of ECG acquisition. For participants who develop the target cancer outcome, event time is defined as the interval from baseline to the earliest cancer diagnosis date. For participants without the target event, follow-up is censored at the earliest date of death, loss to follow-up, or the prespecified study end date. 

We conducted a timepoint-wise PheWAS-style association analysis to evaluate associations between 12-lead ECG waveforms and six continuous gene scores. Each participant contributed 12 ECG leads with 5,000 samples per lead at a sampling rate of 500 Hz, yielding 60,000 lead--timepoint features. For each gene score, we fitted 60,000 linear regression models, with the gene score as the outcome and the ECG amplitude at one lead--timepoint feature as the predictor:
\[
\mathrm{GeneScore}_{i}=\alpha_j+\beta_j\mathrm{ECG}_{ij}+\varepsilon_i,
\]
where $i$ denotes the participant, $j$ denotes the ECG lead--timepoint feature, and $\beta_j$ quantifies the association between ECG amplitude at that feature and the gene score. We used two-sided $t$ tests to calculate $P$ values for the regression coefficients and reported effect estimates ($\beta$), standard errors, correlation coefficients, $t$ statistics, $P$ values, and Benjamini--Hochberg-adjusted $q$ values.

We controlled multiple testing separately for each gene score across its 60,000 tests. We defined FDR-level associations as Benjamini--Hochberg FDR $q<0.05$ and Bonferroni-level associations as $P<0.05/60{,}000=8.33\times10^{-7}$. Manhattan plots display the association strength as $-\log_{10}(P)$ across concatenated ECG features ordered by leads I, II, III, aVR, aVL, aVF, and V1--V6.

To localize association signals within cardiac phases, we detected R peaks from the lead-specific mean ECG signals and extracted beat windows from 300 ms before to 500 ms after each R peak. We used the detected P-wave, QRS-complex, and T-wave boundaries to map association time points to the P wave, QRS complex, T wave, or other phases. We then summarized the number and proportion of FDR- and Bonferroni-level association signals in each phase and identified lead- and phase-specific clusters.

\newpage


\section*{RESOURCE AVAILABILITY}
The authors cannot redistribute individual-level UK Biobank data under the applicable Material Transfer Agreement. Researchers may request access through UK Biobank's controlled-access process (\url{https://www.ukbiobank.ac.uk/use-our-data/apply-for-access/}). This study was conducted under UKB application number 90018.

The Ethical Review Committee of the Chinese Center for Disease Control and Prevention (Beijing, China), the Peking University Institutional Review Board (Beijing, China), and the Oxford Tropical Research Ethics Committee, University of Oxford (UK) approved the study.


PhysioNet provides the MIMIC-IV-ECG Diagnostic Electrocardiogram Matched Subset v1.0 under the Open Data Commons Open Database License v1.0 (\url{https://doi.org/10.13026/4nqg-sb35}). PhysioNet restricts access to the linked MIMIC-IV v2.2 clinical data to credentialed users who complete the required CITI training and sign the data-use agreement (\url{https://doi.org/10.13026/6mm1-ek67}).

The ECG2Gene repository provides the preprocessing and evaluation code (\url{https://github.com/zhangshanwei-mm/ECG2Gene}). The repository contains no participant-level data from controlled-access cohorts. The lead contact will provide additional information needed to reproduce the reported analyses, subject to the applicable data-use agreements.

\section*{ACKNOWLEDGMENTS}

This work was supported by the National Natural Science Foundation of China (Grant No. 82600963). Beijing Nova Program (202604841257), Beijing Municipal Science and Technology Commission (Z251100000725008), Capital’s Funds for Health Improvement and Research (CFH2026-1-4092), Research Project of Peking University in the State Key Laboratory of Vascular Homeostasis and Remodeling (2025-SKLVHR-YCTS-02).

The most important acknowledgment is to the participants in the study and the members of the survey teams in each of the 10 regional centres, as well as to the project development and management teams based at Beijing, Oxford and the 10 regional centres.

The CKB baseline survey and the first re-survey were supported by the Kadoorie Charitable Foundation in Hong Kong. The long-term follow-up has been supported by the Noncommunicable Chronic Diseases-National Science and Technology Major Project (2023ZD0510101, 2023ZD0510100), National Natural Science Foundation of China (82192900, 81390540, 82388102, 91846303, 81941018), the grants (2016YFC0900500) from the National Key R\&D Program of China, and the Chinese Ministry of Science and Technology (2011BAI09B01), and by Wellcome grants to Oxford University (212946/Z/18/Z, 202922/Z/16/Z, 104085/Z/14/Z, 088158/Z/09/Z).




\section*{DECLARATION OF INTERESTS}


The authors declare no competing interests.













\newpage


\bibliography{reference}

\bigskip

\newpage
\clearpage
\appendix
\subsection*{Supplementary Materials}

\setcounter{table}{0}
\setcounter{figure}{0}

\renewcommand{\thetable}{S\arabic{table}}
\renewcommand{\thefigure}{S\arabic{figure}}

\providecommand*{\theHtable}{}
\providecommand*{\theHfigure}{}
\renewcommand*{\theHtable}{supp.table.\arabic{table}}
\renewcommand*{\theHfigure}{supp.figure.\arabic{figure}}


\section*{UK Biobank}
\subsection*{AUROC in the Overall UK Biobank Cohort}

\begin{table}[htbp]
\centering
\caption{Cancer Risk Prediction Performance Using Different XGBoost Inputs}
\label{tab:xgboost_auc_ukb}
\setlength{\tabcolsep}{3pt}
\renewcommand{\arraystretch}{1.05}
\footnotesize
\begin{threeparttable}

\begin{tabularx}{\textwidth}{
@{}
>{\centering\arraybackslash}p{0.075\textwidth}
>{\raggedright\arraybackslash}p{0.38\textwidth}
>{\centering\arraybackslash}X
>{\centering\arraybackslash}X
>{\centering\arraybackslash}X
@{}}
\toprule

\textbf{Group} &
\multicolumn{1}{>{\centering\arraybackslash}p{0.38\textwidth}}{\textbf{XGBoost Input}} &
\thead{Breast Cancer\\AUROC\\(95\% CI)} &
\thead{Ovarian Cancer\\AUROC\\(95\% CI)} &
\thead{Prostate Cancer\\AUROC\\(95\% CI)} \\
\midrule

All & Gene Score + Sex + Age + BMI + Smoke
& $0.820 \pm 0.035$ & $0.719 \pm 0.162$ & $0.810 \pm 0.028$ \\

Female & Age + BMI + Smoke
& $0.641 \pm 0.075$ & $0.405 \pm 0.219$ & -- \\

Female & Gene Score
& $0.516 \pm 0.069$ & $0.542 \pm 0.278$ & -- \\

Female & Gene Score+ Age
& $0.643 \pm 0.065$ & $0.598 \pm 0.251$ & -- \\

Female & Gene Score+ BMI
& $0.593 \pm 0.069$ & $0.495 \pm 0.282$ & -- \\

Female & Gene Score+ Smoke
& $0.539 \pm 0.072$ & $0.511 \pm 0.268$ & -- \\

Female & Gene Score+ Age + BMI
& $0.669 \pm 0.066$ & $0.604 \pm 0.248$ & -- \\

Female & Gene Score+ Age + Smoke
& $0.649 \pm 0.068$ & $0.633 \pm 0.241$ & -- \\

Female & Gene Score+ BMI + Smoke
& $0.593 \pm 0.073$ & $0.507 \pm 0.304$ & -- \\

Female & Gene Score+ Age + BMI + Smoke
& $0.683 \pm 0.065$ & $0.580 \pm 0.264$ & -- \\

Male & Age + BMI + Smoke
& -- & -- & $0.616 \pm 0.054$ \\

Male & Gene Score
& -- & -- & $0.581 \pm 0.055$ \\

Male & Gene Score + Age
& -- & -- & $0.656 \pm 0.054$ \\

Male & Gene Score + BMI
& -- & -- & $0.609 \pm 0.053$ \\

Male & Gene Score + Smoke
& -- & -- & $0.576 \pm 0.055$ \\

Male & Gene Score + Age + BMI
& -- & -- & $0.660 \pm 0.057$ \\

Male & Gene Score + Age + Smoke
& -- & -- & $0.645 \pm 0.054$ \\

Male & Gene Score + BMI + Smoke
& -- & -- & $0.604 \pm 0.055$ \\

Male & Gene Score + Age + BMI + Smoke
& -- & -- & $0.650 \pm 0.056$ \\

\bottomrule
\end{tabularx}
\end{threeparttable}
\end{table}

\subsection*{AUROC Among White Participants in the UK Biobank Cohort}

\begin{table}[H]
\centering
\caption{Cancer Risk Prediction Performance Using Different XGBoost Inputs Among White Participants}
\label{tab:ukb_auc_white}
\setlength{\tabcolsep}{3pt}
\renewcommand{\arraystretch}{1.05}
\footnotesize
\begin{threeparttable}
\begin{tabularx}{\textwidth}{
@{}
>{\centering\arraybackslash}p{0.075\textwidth}
>{\raggedright\arraybackslash}p{0.38\textwidth}
>{\centering\arraybackslash}X
>{\centering\arraybackslash}X
>{\centering\arraybackslash}X
@{}}
\toprule
\textbf{Group} &
\multicolumn{1}{>{\centering\arraybackslash}p{0.38\textwidth}}{\textbf{XGBoost Input}} &
\thead{Breast Cancer\\AUROC} &
\thead{Ovarian Cancer\\AUROC} &
\thead{Prostate Cancer\\AUROC} \\
\midrule
All & Gene Score + Sex + Age + BMI + Smoke
& $0.818 \pm 0.035$ & $0.719 \pm 0.166$ & $0.814 \pm 0.027$ \\
Female & Age + BMI + Smoke
& $0.637 \pm 0.076$ & $0.404 \pm 0.214$ & -- \\
Female & Gene Score
& $0.509 \pm 0.071$ & $0.542 \pm 0.283$ & -- \\
Female & Gene Score + Age
& $0.639 \pm 0.068$ & $0.599 \pm 0.257$ & -- \\
Female & Gene Score + BMI
& $0.591 \pm 0.073$ & $0.495 \pm 0.287$ & -- \\
Female & Gene Score + Smoke
& $0.532 \pm 0.075$ & $0.511 \pm 0.272$ & -- \\
Female & Gene Score + Age + BMI
& $0.666 \pm 0.068$ & $0.605 \pm 0.258$ & -- \\
Female & Gene Score + Age + Smoke
& $0.645 \pm 0.070$ & $0.633 \pm 0.249$ & -- \\
Female & Gene Score + BMI + Smoke
& $0.592 \pm 0.076$ & $0.506 \pm 0.305$ & -- \\
Female & Gene Score + Age + BMI + Smoke
& $0.680 \pm 0.066$ & $0.581 \pm 0.284$ & -- \\
Male & Age + BMI + Smoke
& -- & -- & $0.619 \pm 0.052$ \\
Male & Gene Score
& -- & -- & $0.590 \pm 0.056$ \\
Male & Gene Score + Age
& -- & -- & $0.664 \pm 0.053$ \\
Male & Gene Score + BMI
& -- & -- & $0.614 \pm 0.055$ \\
Male & Gene Score + Smoke
& -- & -- & $0.584 \pm 0.053$ \\
Male & Gene Score + Age + BMI
& -- & -- & $0.668 \pm 0.056$ \\
Male & Gene Score + Age + Smoke
& -- & -- & $0.652 \pm 0.054$ \\
Male & Gene Score + BMI + Smoke
& -- & -- & $0.609 \pm 0.054$ \\
Male & Gene Score + Age + BMI + Smoke
& -- & -- & $0.657 \pm 0.056$ \\
\bottomrule
\end{tabularx}
\end{threeparttable}
\end{table}


\subsection*{Cancer Risk Prediction AUROC Using Pathogenic Variant Inputs Among UK Biobank Participants}

\begin{table}[H]
\centering
\caption{Cancer Risk Prediction Performance Using Different Pathogenic Variants Inputs}
\label{tab:ukb_pv_auc}
\setlength{\tabcolsep}{3pt}
\renewcommand{\arraystretch}{1.05}
\footnotesize

\begin{threeparttable}
\begin{tabularx}{\textwidth}{
@{}
>{\centering\arraybackslash}p{0.08\textwidth}
>{\raggedright\arraybackslash}p{0.38\textwidth}
>{\centering\arraybackslash}X
>{\centering\arraybackslash}X
>{\centering\arraybackslash}X
@{}}
\toprule
\textbf{Group} &
\textbf{XGBoost Input} &
\thead{Breast Cancer\\AUROC} &
\thead{Ovarian Cancer\\AUROC} &
\thead{Prostate Cancer\\AUROC} \\
\midrule

All & PV + Sex + Age + BMI + Smoke
& $0.811 \pm 0.037$ & $0.867 \pm 0.096$ & $0.798 \pm 0.026$ \\

Female & Age + BMI + Smoke
& $0.641 \pm 0.075$ & $0.405 \pm 0.219$ & -- \\

Female & PV + Age
& $0.624 \pm 0.072$ & $0.693 \pm 0.238$ & -- \\

Female & PV + BMI
& $0.596 \pm 0.080$ & $0.465 \pm 0.286$ & -- \\

Female & PV + Smoke
& $0.506 \pm 0.062$ & $0.451 \pm 0.250$ & -- \\

Female & PV + Age + BMI
& $0.662 \pm 0.071$ & $0.600 \pm 0.262$ & -- \\

Female & PV + Age + Smoke
& $0.638 \pm 0.069$ & $0.687 \pm 0.241$ & -- \\

Female & PV + BMI + Smoke
& $0.589 \pm 0.076$ & $0.587 \pm 0.182$ & -- \\

Female & PV + Age + BMI + Smoke
& $0.648 \pm 0.074$ & $0.695 \pm 0.261$ & -- \\

Male & Age + BMI + Smoke
& -- & -- & $0.616 \pm 0.054$ \\

Male & PV + Age
& -- & -- & $0.608 \pm 0.050$ \\

Male & PV + BMI
& -- & -- & $0.552 \pm 0.052$ \\

Male & PV + Smoke
& -- & -- & $0.501 \pm 0.048$ \\

Male & PV + Age + BMI
& -- & -- & $0.648 \pm 0.056$ \\

Male & PV + Age + Smoke
& -- & -- & $0.590 \pm 0.052$ \\

Male & PV + BMI + Smoke
& -- & -- & $0.542 \pm 0.056$ \\

Male & PV + Age + BMI + Smoke
& -- & -- & $0.625 \pm 0.053$ \\

\bottomrule
\end{tabularx}
\begin{tablenotes}[flushleft]
\scriptsize
\item[] PV indicates positive carrier status for pathogenic variants in the six input genes.
\end{tablenotes}

\end{threeparttable}
\end{table}

\subsection*{Cancer Risk Prediction AUROC Using Pathogenic Variant Inputs Among White UK Biobank Participants}
\begin{table}[H]
\centering
\caption{Cancer Risk Prediction Performance Using Different XGBoost Inputs Among White Participants}
\label{tab:ukb_auc_white_pv}
\setlength{\tabcolsep}{3pt}
\renewcommand{\arraystretch}{1.05}
\footnotesize
\begin{threeparttable}
\begin{tabularx}{\textwidth}{
@{}
>{\centering\arraybackslash}p{0.075\textwidth}
>{\raggedright\arraybackslash}p{0.38\textwidth}
>{\centering\arraybackslash}X
>{\centering\arraybackslash}X
>{\centering\arraybackslash}X
@{}}
\toprule
\textbf{Group} &
\multicolumn{1}{>{\centering\arraybackslash}p{0.38\textwidth}}{\textbf{XGBoost Input}} &
\thead{Breast Cancer\\AUROC} &
\thead{Ovarian Cancer\\AUROC} &
\thead{Prostate Cancer\\AUROC} \\
\midrule
All & PV + Sex + Age + BMI + Smoke
& $0.811 \pm 0.038$ & $0.869 \pm 0.097$ & $0.798 \pm 0.026$ \\
Female & Age + BMI
& $0.636 \pm 0.077$ & $0.443 \pm 0.272$ & -- \\
Female & Age + BMI + Smoke
& $0.637 \pm 0.076$ & $0.404 \pm 0.214$ & -- \\
Female & PV
& $0.500 \pm 0.000$ & $0.500 \pm 0.000$ & -- \\
Female & PV + Age
& $0.622 \pm 0.072$ & $0.695 \pm 0.237$ & -- \\
Female & PV + BMI
& $0.595 \pm 0.081$ & $0.465 \pm 0.287$ & -- \\
Female & PV + Smoke
& $0.505 \pm 0.064$ & $0.453 \pm 0.250$ & -- \\
Female & PV + Age + BMI
& $0.660 \pm 0.069$ & $0.602 \pm 0.260$ & -- \\
Female & PV + Age + Smoke
& $0.635 \pm 0.068$ & $0.689 \pm 0.232$ & -- \\
Female & PV + BMI + Smoke
& $0.588 \pm 0.079$ & $0.585 \pm 0.174$ & -- \\
Female & PV + Age + BMI + Smoke
& $0.646 \pm 0.075$ & $0.697 \pm 0.262$ & -- \\
Male & Age + BMI
& -- & -- & $0.620 \pm 0.052$ \\
Male & Age + BMI + Smoke
& -- & -- & $0.619 \pm 0.052$ \\
Male & PV
& -- & -- & $0.500 \pm 0.000$ \\
Male & PV + Age
& -- & -- & $0.611 \pm 0.048$ \\
Male & PV + BMI
& -- & -- & $0.551 \pm 0.054$ \\
Male & PV + Smoke
& -- & -- & $0.501 \pm 0.050$ \\
Male & PV + Age + BMI
& -- & -- & $0.650 \pm 0.055$ \\
Male & PV + Age + Smoke
& -- & -- & $0.594 \pm 0.050$ \\
Male & PV + BMI + Smoke
& -- & -- & $0.542 \pm 0.058$ \\
Male & PV + Age + BMI + Smoke
& -- & -- & $0.627 \pm 0.053$ \\
\bottomrule
\end{tabularx}

\begin{tablenotes}[flushleft]
\scriptsize
\item[] PV indicates positive carrier status for pathogenic variants in the six input genes.
\end{tablenotes}

\end{threeparttable}
\end{table}

\subsection*{AUROC in UK Biobank Participants With CVD}
\begin{table}[H]
\centering
\caption{Cancer Risk Prediction Performance Using Different XGBoost Inputs Among Participants with Cardiovascular Disease}
\label{tab:ukb_auc_cvd}
\setlength{\tabcolsep}{3pt}
\renewcommand{\arraystretch}{1.05}
\footnotesize
\begin{threeparttable}
\begin{tabularx}{\textwidth}{
@{}
>{\centering\arraybackslash}p{0.075\textwidth}
>{\raggedright\arraybackslash}p{0.38\textwidth}
>{\centering\arraybackslash}X
>{\centering\arraybackslash}X
>{\centering\arraybackslash}X
@{}}
\toprule
\textbf{Group} &
\multicolumn{1}{>{\centering\arraybackslash}p{0.38\textwidth}}{\textbf{XGBoost Input}} &
\thead{Breast Cancer\\AUROC} &
\thead{Ovarian Cancer\\AUROC} &
\thead{Prostate Cancer\\AUROC} \\
\midrule
All & Gene Score + Sex + Age + BMI + Smoke
& $0.899 \pm 0.062$ & NA & $0.728 \pm 0.049$ \\
Female & Age + BMI
& $0.663 \pm 0.179$ & NA & -- \\
Female & Age + BMI + Smoke
& $0.669 \pm 0.186$ & NA & -- \\
Female & Gene Score
& $0.710 \pm 0.163$ & NA & -- \\
Female & Gene Score + Age
& $0.749 \pm 0.137$ & NA & -- \\
Female & Gene Score + BMI
& $0.502 \pm 0.197$ & NA & -- \\
Female & Gene Score + Smoke
& $0.508 \pm 0.234$ & NA & -- \\
Female & Gene Score + Age + BMI
& $0.737 \pm 0.148$ & NA & -- \\
Female & Gene Score + Age + Smoke
& $0.749 \pm 0.130$ & NA & -- \\
Female & Gene Score + BMI + Smoke
& $0.534 \pm 0.205$ & NA & -- \\
Female & Gene Score + Age + BMI + Smoke
& $0.710 \pm 0.173$ & NA & -- \\
Male & Age + BMI
& -- & -- & $0.601 \pm 0.082$ \\
Male & Age + BMI + Smoke
& -- & -- & $0.569 \pm 0.075$ \\
Male & Gene Score
& -- & -- & $0.527 \pm 0.088$ \\
Male & Gene Score + Age
& -- & -- & $0.597 \pm 0.087$ \\
Male & Gene Score + BMI
& -- & -- & $0.565 \pm 0.082$ \\
Male & Gene Score + Smoke
& -- & -- & $0.505 \pm 0.087$ \\
Male & Gene Score + Age + BMI
& -- & -- & $0.607 \pm 0.082$ \\
Male & Gene Score + Age + Smoke
& -- & -- & $0.562 \pm 0.083$ \\
Male & Gene Score + BMI + Smoke
& -- & -- & $0.528 \pm 0.085$ \\
Male & Gene Score + Age + BMI + Smoke
& -- & -- & $0.575 \pm 0.085$ \\
\bottomrule
\end{tabularx}
\begin{tablenotes}[flushleft]
\footnotesize
\item NA indicates that AUROC could not be estimated because no ovarian cancer cases were observed in this subgroup.
\end{tablenotes}
\end{threeparttable}
\end{table}

\subsection*{AUROC in UK Biobank Participants Without CVD}

\begin{table}[H]
\centering
\caption{Cancer Risk Prediction Performance Using Different XGBoost Inputs Among Participants without Cardiovascular Disease}
\label{tab:ukb_auc_no_cvd}
\setlength{\tabcolsep}{3pt}
\renewcommand{\arraystretch}{1.05}
\footnotesize
\begin{threeparttable}
\begin{tabularx}{\textwidth}{
@{}
>{\centering\arraybackslash}p{0.075\textwidth}
>{\raggedright\arraybackslash}p{0.38\textwidth}
>{\centering\arraybackslash}X
>{\centering\arraybackslash}X
>{\centering\arraybackslash}X
@{}}
\toprule
\textbf{Group} &
\multicolumn{1}{>{\centering\arraybackslash}p{0.38\textwidth}}{\textbf{XGBoost Input}} &
\thead{Breast Cancer\\AUROC} &
\thead{Ovarian Cancer\\AUROC} &
\thead{Prostate Cancer\\AUROC} \\
\midrule
All & Gene Score + Sex + Age + BMI + Smoke
& $0.800 \pm 0.040$ & $0.706 \pm 0.166$ & $0.831 \pm 0.036$ \\
Female & Age + BMI
& $0.634 \pm 0.086$ & $0.447 \pm 0.264$ & -- \\
Female & Age + BMI + Smoke
& $0.634 \pm 0.079$ & $0.410 \pm 0.227$ & -- \\
Female & Gene Score
& $0.676 \pm 0.073$ & $0.576 \pm 0.247$ & -- \\
Female & Gene Score + Age
& $0.619 \pm 0.075$ & $0.590 \pm 0.250$ & -- \\
Female & Gene Score + BMI
& $0.612 \pm 0.075$ & $0.495 \pm 0.274$ & -- \\
Female & Gene Score + Smoke
& $0.545 \pm 0.078$ & $0.509 \pm 0.282$ & -- \\
Female & Gene Score + Age + BMI
& $0.654 \pm 0.076$ & $0.595 \pm 0.254$ & -- \\
Female & Gene Score + Age + Smoke
& $0.627 \pm 0.077$ & $0.627 \pm 0.244$ & -- \\
Female & Gene Score + BMI + Smoke
& $0.606 \pm 0.082$ & $0.509 \pm 0.296$ & -- \\
Female & Gene Score + Age + BMI + Smoke
& $0.676 \pm 0.074$ & $0.576 \pm 0.286$ & -- \\
Male & Age + BMI
& -- & -- & $0.603 \pm 0.076$ \\
Male & Age + BMI + Smoke
& -- & -- & $0.614 \pm 0.072$ \\
Male & Gene Score
& -- & -- & $0.608 \pm 0.072$ \\
Male & Gene Score + Age
& -- & -- & $0.669 \pm 0.069$ \\
Male & Gene Score + BMI
& -- & -- & $0.622 \pm 0.072$ \\
Male & Gene Score + Smoke
& -- & -- & $0.612 \pm 0.076$ \\
Male & Gene Score + Age + BMI
& -- & -- & $0.668 \pm 0.076$ \\
Male & Gene Score + Age + Smoke
& -- & -- & $0.668 \pm 0.073$ \\
Male & Gene Score + BMI + Smoke
& -- & -- & $0.636 \pm 0.071$ \\
Male & Gene Score + Age + BMI + Smoke
& -- & -- & $0.664 \pm 0.077$ \\
\bottomrule
\end{tabularx}
\end{threeparttable}
\end{table}

\subsection*{Kaplan--Meier risk stratification by ECG-derived gene scores}
\subsubsection*{Female UK Biobank participants with breast/ovarian cancer outcomes}
\begin{figure}[H] %
\centerline{\includegraphics[width=1\textwidth]{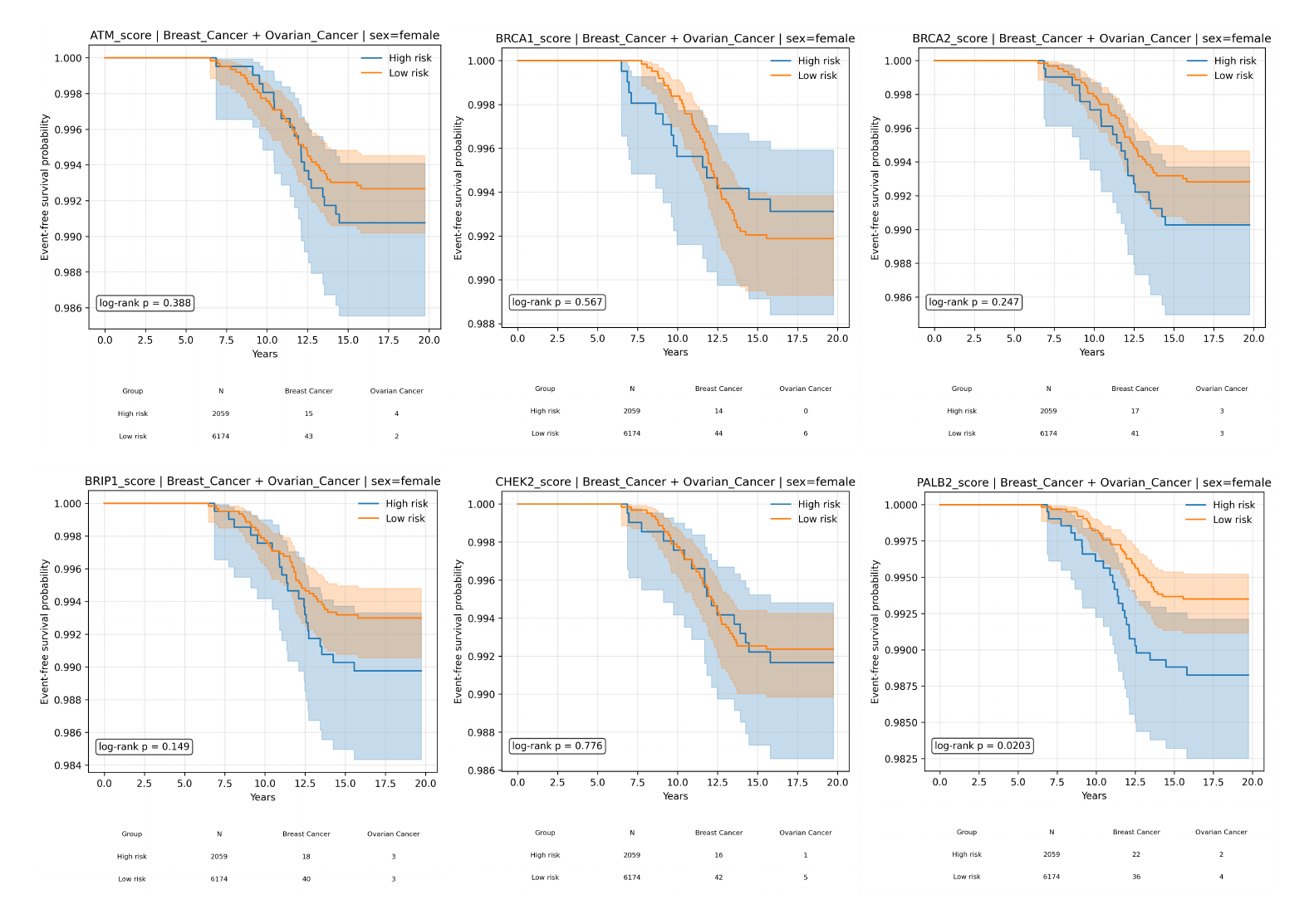}}
\caption{KM curves for breast cancer and ovarian cancer outcomes stratified by six gene prediction scores in female participants from UK Biobank.}
\label{km-ukb-female}
\end{figure}

\subsubsection*{Male UK Biobank participants with prostate cancer outcomes}
\begin{figure}[H] %
\centerline{\includegraphics[width=1\textwidth]{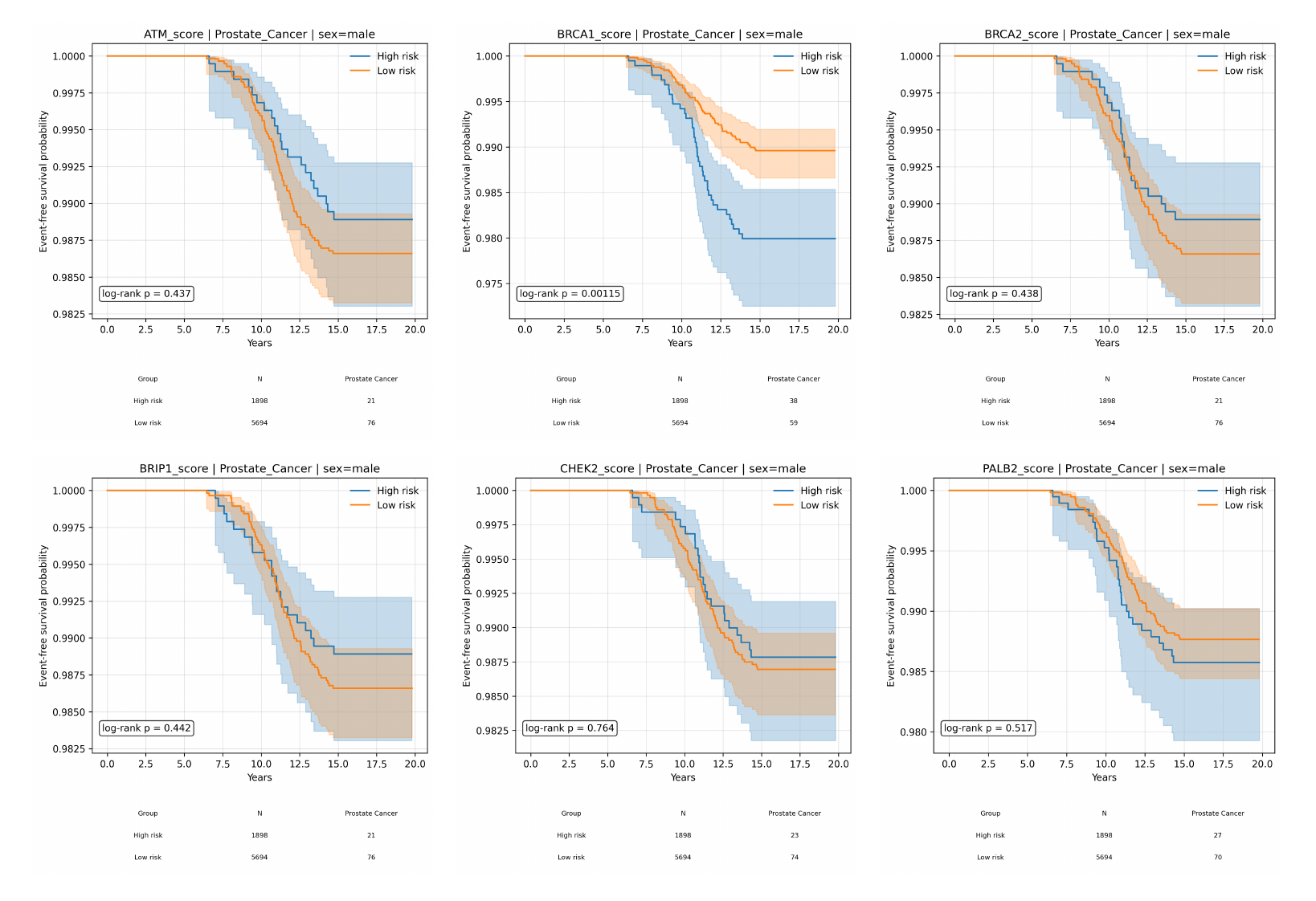}}
\caption{
KM curves for prostate cancer outcomes stratified by six gene prediction scores in male participants from UK Biobank.}
\label{km-ukb-male}
\end{figure}

\section*{MIMIC}

\subsection*{MIMIC Female Participants: White vs. Non-White}

\begin{table}[H]
\centering
\caption{MIMIC-IV-ECG Female Breast Cancer Prediction AUROC in Overall White and Non-white Groups}
\label{tab:mimic_breast_auc_race}
\setlength{\tabcolsep}{3pt}
\renewcommand{\arraystretch}{1.05}
\footnotesize
\begin{threeparttable}
\begin{tabularx}{\textwidth}{
@{}
>{\centering\arraybackslash}p{0.075\textwidth}
>{\raggedright\arraybackslash}p{0.40\textwidth}
>{\centering\arraybackslash}X
>{\centering\arraybackslash}X
>{\centering\arraybackslash}X
@{}}
\toprule
\textbf{Group} &
\multicolumn{1}{>{\centering\arraybackslash}p{0.40\textwidth}}{\textbf{XGBoost Input}} &
\thead{Overall\\AUROC} &
\thead{White\\AUROC} &
\thead{Non-white\\AUROC} \\
\midrule
Female & Age + BMI + Smoke
& $0.488$ & $0.493$ & $0.489$ \\
Female & Gene Score
& $0.491$ & $0.492$ & $0.498$ \\
Female & Gene Score + Age
& $0.474$ & $0.500$ & $0.461$ \\
Female & Gene Score + BMI
& $0.513$ & $0.501$ & $0.529$ \\
Female & Gene Score + Smoke
& $0.489$ & $0.493$ & $0.489$ \\
Female & Gene Score + Age + BMI
& $0.489$ & $0.493$ & $0.491$ \\
Female & Gene Score + Age + Smoke
& $0.471$ & $0.497$ & $0.458$ \\
Female & Gene Score + BMI + Smoke
& $0.522$ & $0.505$ & $0.547$ \\
Female & Gene Score + Age + BMI + Smoke
& $0.491$ & $0.491$ & $0.499$ \\
\bottomrule
\end{tabularx}
\end{threeparttable}
\end{table}

\begin{table}[H]
\centering
\caption{MIMIC-IV-ECG Female Ovarian Cancer Prediction AUROC in Overall White and Non-white Groups}
\label{tab:mimic_ovarian_auc_race}
\setlength{\tabcolsep}{3pt}
\renewcommand{\arraystretch}{1.05}
\footnotesize
\begin{threeparttable}
\begin{tabularx}{\textwidth}{
@{}
>{\centering\arraybackslash}p{0.075\textwidth}
>{\raggedright\arraybackslash}p{0.40\textwidth}
>{\centering\arraybackslash}X
>{\centering\arraybackslash}X
>{\centering\arraybackslash}X
@{}}
\toprule
\textbf{Group} &
\multicolumn{1}{>{\centering\arraybackslash}p{0.40\textwidth}}{\textbf{XGBoost Input}} &
\thead{Overall\\AUROC} &
\thead{White\\AUROC} &
\thead{Non-white\\AUROC} \\
\midrule
Female & Age + BMI + Smoke
& $0.466$ & $0.476$ & $0.502$ \\
Female & Gene Score
& $0.528$ & $0.530$ & $0.541$ \\
Female & Gene Score + Age
& $0.517$ & $0.540$ & $0.553$ \\
Female & Gene Score + BMI
& $0.535$ & $0.532$ & $0.563$ \\
Female & Gene Score + Smoke
& $0.537$ & $0.538$ & $0.560$ \\
Female & Gene Score + Age + BMI
& $0.528$ & $0.543$ & $0.557$ \\
Female & Gene Score + Age + Smoke
& $0.524$ & $0.544$ & $0.550$ \\
Female & Gene Score + BMI + Smoke
& $0.538$ & $0.539$ & $0.552$ \\
Female & Gene Score + Age + BMI + Smoke
& $0.522$ & $0.539$ & $0.532$ \\
\bottomrule
\end{tabularx}
\end{threeparttable}
\end{table}

\subsection*{MIMIC Male Participants: White vs. Non-White}

\begin{table}[H]
\centering
\caption{MIMIC-IV-ECG Male Prostate Cancer Prediction AUROC in Overall White and Non-white Groups}
\label{tab:mimic_prostate_auc_race}
\setlength{\tabcolsep}{3pt}
\renewcommand{\arraystretch}{1.05}
\footnotesize
\begin{threeparttable}
\begin{tabularx}{\textwidth}{
@{}
>{\centering\arraybackslash}p{0.075\textwidth}
>{\raggedright\arraybackslash}p{0.40\textwidth}
>{\centering\arraybackslash}X
>{\centering\arraybackslash}X
>{\centering\arraybackslash}X
@{}}
\toprule
\textbf{Group} &
\multicolumn{1}{>{\centering\arraybackslash}p{0.40\textwidth}}{\textbf{XGBoost Input}} &
\thead{Overall\\AUROC} &
\thead{White\\AUROC} &
\thead{Non-white\\AUROC} \\
\midrule
Male & Age + BMI + Smoke
& $0.550$ & $0.514$ & $0.605$ \\
Male & Gene Score
& $0.514$ & $0.522$ & $0.499$ \\
Male & Gene Score + Age
& $0.579$ & $0.559$ & $0.607$ \\
Male & Gene Score + BMI
& $0.514$ & $0.520$ & $0.514$ \\
Male & Gene Score + Smoke
& $0.491$ & $0.480$ & $0.499$ \\
Male & Gene Score + Age + BMI
& $0.558$ & $0.545$ & $0.589$ \\
Male & Gene Score + Age + Smoke
& $0.553$ & $0.524$ & $0.591$ \\
Male & Gene Score + BMI + Smoke
& $0.483$ & $0.470$ & $0.507$ \\
Male & Gene Score + Age + BMI + Smoke
& $0.531$ & $0.507$ & $0.572$ \\
\bottomrule
\end{tabularx}
\end{threeparttable}
\end{table}

\subsection*{AUROC in MIMIC Participants Without CVD}

\begin{table}[H]
\centering
\caption{AUROC in MIMIC Participants Without CVD}
\label{tab:minic_auc_without_cvd}
\setlength{\tabcolsep}{3pt}
\renewcommand{\arraystretch}{1.05}
\footnotesize

\begin{threeparttable}
\begin{tabularx}{\textwidth}{
@{}
>{\centering\arraybackslash}p{0.08\textwidth}
>{\centering\arraybackslash}p{0.38\textwidth}
>{\centering\arraybackslash}X
>{\centering\arraybackslash}X
>{\centering\arraybackslash}X
@{}}
\toprule
\textbf{Group} &
\textbf{XGBoost Input} &
\thead{Breast Cancer\\AUROC (95\% CI)} &
\thead{Ovarian Cancer\\AUROC (95\% CI)} &
\thead{Prostate Cancer\\AUROC (95\% CI)} \\
\midrule

All & Gene Score + Sex + Age + BMI + Smoke
& $0.679 \pm 0.010$ & $0.670 \pm 0.021$ & $0.837 \pm 0.006$ \\

Female & Age + BMI
& $0.457 \pm 0.014$ & $0.411 \pm 0.029$ & -- \\

Female & Age + BMI + Smoke
& $0.453 \pm 0.015$ & $0.408 \pm 0.028$ & -- \\

Female & Gene Score
& $0.514 \pm 0.014$ & $0.532 \pm 0.027$ & -- \\

Female & Gene Score + Age
& $0.458 \pm 0.012$ & $0.470 \pm 0.027$ & -- \\

Female & Gene Score + BMI
& $0.507 \pm 0.016$ & $0.527 \pm 0.032$ & -- \\

Female & Gene Score + Smoke
& $0.518 \pm 0.014$ & $0.540 \pm 0.026$ & -- \\

Female & Gene Score + Age + BMI
& $0.465 \pm 0.015$ & $0.480 \pm 0.030$ & -- \\

Female & Gene Score + Age + Smoke
& $0.449 \pm 0.012$ & $0.487 \pm 0.027$ & -- \\

Female & Gene Score + BMI + Smoke
& $0.510 \pm 0.016$ & $0.533 \pm 0.030$ & -- \\

Female & Gene Score + Age + BMI + Smoke
& $0.470 \pm 0.014$ & $0.473 \pm 0.031$ & -- \\

Male & Age + BMI
& -- & -- & $0.616 \pm 0.014$ \\

Male & Age + BMI + Smoke
& -- & -- & $0.610 \pm 0.014$ \\

Male & Gene Score
& -- & -- & $0.548 \pm 0.011$ \\

Male & Gene Score + Age
& -- & -- & $0.639 \pm 0.010$ \\

Male & Gene Score + BMI
& -- & -- & $0.553 \pm 0.014$ \\

Male & Gene Score + Smoke
& -- & -- & $0.541 \pm 0.011$ \\

Male & Gene Score + Age + BMI
& -- & -- & $0.618 \pm 0.014$ \\

Male & Gene Score + Age + Smoke
& -- & -- & $0.621 \pm 0.011$ \\

Male & Gene Score + BMI + Smoke
& -- & -- & $0.545 \pm 0.014$ \\

Male & Gene Score + Age + BMI + Smoke
& -- & -- & $0.608 \pm 0.014$ \\

\bottomrule
\end{tabularx}
\end{threeparttable}
\end{table}

\subsection*{AUROC in MIMIC Participants With CVD}

\begin{table}[H]
\centering
\caption{AUROC in MIMIC Participants With CVD}
\label{tab:minic_auc_with_cvd}
\setlength{\tabcolsep}{3pt}
\renewcommand{\arraystretch}{1.05}
\footnotesize

\begin{threeparttable}
\begin{tabularx}{\textwidth}{
@{}
>{\centering\arraybackslash}p{0.08\textwidth}
>{\centering\arraybackslash}p{0.38\textwidth}
>{\centering\arraybackslash}X
>{\centering\arraybackslash}X
>{\centering\arraybackslash}X
@{}}
\toprule
\textbf{Group} &
\textbf{XGBoost Input} &
\thead{Breast Cancer\\AUROC (95\% CI)} &
\thead{Ovarian Cancer\\AUROC (95\% CI)} &
\thead{Prostate Cancer\\AUROC (95\% CI)} \\
\midrule

All & Gene Score + Sex + Age + BMI + Smoke
& $0.735 \pm 0.008$ & $0.739 \pm 0.021$ & $0.761 \pm 0.006$ \\

Female & Age + BMI
& $0.490 \pm 0.014$ & $0.443 \pm 0.038$ & -- \\

Female & Age + BMI + Smoke
& $0.488 \pm 0.013$ & $0.478 \pm 0.037$ & -- \\

Female & Gene Score
& $0.460 \pm 0.013$ & $0.513 \pm 0.032$ & -- \\

Female & Gene Score + Age
& $0.457 \pm 0.012$ & $0.527 \pm 0.033$ & -- \\

Female & Gene Score + BMI
& $0.512 \pm 0.014$ & $0.535 \pm 0.035$ & -- \\

Female & Gene Score + Smoke
& $0.455 \pm 0.012$ & $0.511 \pm 0.033$ & -- \\

Female & Gene Score + Age + BMI
& $0.476 \pm 0.014$ & $0.527 \pm 0.038$ & -- \\

Female & Gene Score + Age + Smoke
& $0.457 \pm 0.012$ & $0.526 \pm 0.031$ & -- \\

Female & Gene Score + BMI + Smoke
& $0.526 \pm 0.014$ & $0.530 \pm 0.036$ & -- \\

Female & Gene Score + Age + BMI + Smoke
& $0.478 \pm 0.013$ & $0.536 \pm 0.035$ & -- \\

Male & Age + BMI
& -- & -- & $0.563 \pm 0.012$ \\

Male & Age + BMI + Smoke
& -- & -- & $0.540 \pm 0.012$ \\

Male & Gene Score
& -- & -- & $0.496 \pm 0.010$ \\

Male & Gene Score + Age
& -- & -- & $0.550 \pm 0.010$ \\

Male & Gene Score + BMI
& -- & -- & $0.497 \pm 0.012$ \\

Male & Gene Score + Smoke
& -- & -- & $0.481 \pm 0.010$ \\

Male & Gene Score + Age + BMI
& -- & -- & $0.536 \pm 0.011$ \\

Male & Gene Score + Age + Smoke
& -- & -- & $0.535 \pm 0.010$ \\

Male & Gene Score + BMI + Smoke
& -- & -- & $0.478 \pm 0.012$ \\

Male & Gene Score + Age + BMI + Smoke
& -- & -- & $0.516 \pm 0.012$ \\

\bottomrule
\end{tabularx}
\end{threeparttable}
\end{table}

\section*{China Kadoorie Biobank}
\subsection*{AUROC in the Overall CKB Cohort}

\begin{table}[H]
\centering
\caption{Cancer Risk Prediction Performance Using Different XGBoost Inputs}
\label{tab:ckb_auc}
\setlength{\tabcolsep}{3pt}
\renewcommand{\arraystretch}{1.05}
\footnotesize

\begin{threeparttable}
\begin{tabularx}{\textwidth}{
@{}
>{\centering\arraybackslash}p{0.08\textwidth}
>{\centering\arraybackslash}p{0.38\textwidth}
>{\centering\arraybackslash}X
>{\centering\arraybackslash}X
>{\centering\arraybackslash}X
@{}}
\toprule
\textbf{Group} &
\textbf{XGBoost Input} &
\thead{Breast Cancer\\AUROC} &
\thead{Ovarian Cancer\\AUROC} &
\thead{Prostate Cancer\\AUROC} \\
\midrule      All & Gene Score + Sex + Age + BMI + Smoke & $0.546 \pm 0.051$ & $0.455 \pm 0.143$ & $0.682 \pm 0.079$ \\
      Female & Age + BMI + Smoke & $0.536 \pm 0.058$ & $0.407 \pm 0.119$ & -- \\
      Female & Gene Score & $0.511 \pm 0.055$ & $0.483 \pm 0.138$ & -- \\
      Female & Gene Score + Age & $0.536 \pm 0.053$ & $0.443 \pm 0.130$ & -- \\
      Female & Gene Score + BMI & $0.492 \pm 0.055$ & $0.488 \pm 0.144$ & -- \\
      Female & Gene Score + Smoke & $0.507 \pm 0.055$ & $0.493 \pm 0.137$ & -- \\
      Female & Gene Score + Age + BMI & $0.524 \pm 0.052$ & $0.431 \pm 0.130$ & -- \\
      Female & Gene Score + Age + Smoke & $0.531 \pm 0.053$ & $0.473 \pm 0.139$ & -- \\
      Female & Gene Score + BMI + Smoke & $0.489 \pm 0.056$ & $0.497 \pm 0.136$ & -- \\
      Female & Gene Score + Age + BMI + Smoke & $0.526 \pm 0.052$ & $0.449 \pm 0.138$ & -- \\
      Male & Age + BMI + Smoke & -- & -- & $0.638 \pm 0.078$ \\
      Male & Gene Score & -- & -- & $0.477 \pm 0.096$ \\
      Male & Gene Score + Age & -- & -- & $0.709 \pm 0.073$ \\
      Male & Gene Score + BMI & -- & -- & $0.493 \pm 0.096$ \\
      Male & Gene Score + Smoke & -- & -- & $0.437 \pm 0.095$ \\
      Male & Gene Score + Age + BMI & -- & -- & $0.676 \pm 0.083$ \\
      Male & Gene Score + Age + Smoke & -- & -- & $0.664 \pm 0.087$ \\
      Male & Gene Score + BMI + Smoke & -- & -- & $0.421 \pm 0.085$ \\
      Male & Gene Score + Age + BMI + Smoke & -- & -- & $0.638 \pm 0.083$ \\
\bottomrule
\end{tabularx}
\end{threeparttable}
\end{table}

\subsection*{AUROC in CKB Participants Without CVD}

\begin{table}[H]
\centering
\caption{AUROC in CKB Participants Without CVD}
\label{tab:ckb_auc_without_cvd}
\setlength{\tabcolsep}{3pt}
\renewcommand{\arraystretch}{1.05}
\footnotesize

\begin{threeparttable}
\begin{tabularx}{\textwidth}{
@{}
>{\centering\arraybackslash}p{0.08\textwidth}
>{\centering\arraybackslash}p{0.38\textwidth}
>{\centering\arraybackslash}X
>{\centering\arraybackslash}X
>{\centering\arraybackslash}X
@{}}
\toprule
\textbf{Group} &
\textbf{XGBoost Input} &
\thead{Breast Cancer\\AUROC (95\% CI)} &
\thead{Ovarian Cancer\\AUROC (95\% CI)} &
\thead{Prostate Cancer\\AUROC (95\% CI)} \\
\midrule

      All & Gene Score + Sex + Age + BMI + Smoke & $0.561 \pm 0.051$ & $0.448 \pm 0.142$ & $0.683 \pm 0.087$ \\
      Female & Age + BMI + Smoke & $0.553 \pm 0.060$ & $0.395 \pm 0.117$ & -- \\
      Female & Gene Score & $0.533 \pm 0.057$ & $0.484 \pm 0.137$ & -- \\
      Female & Gene Score + Age & $0.558 \pm 0.053$ & $0.434 \pm 0.129$ & -- \\
      Female & Gene Score + BMI & $0.502 \pm 0.057$ & $0.489 \pm 0.143$ & -- \\
      Female & Gene Score + Smoke & $0.528 \pm 0.058$ & $0.494 \pm 0.136$ & -- \\
      Female & Gene Score + Age + BMI & $0.541 \pm 0.052$ & $0.422 \pm 0.130$ & -- \\
      Female & Gene Score + Age + Smoke & $0.550 \pm 0.052$ & $0.466 \pm 0.138$ & -- \\
      Female & Gene Score + BMI + Smoke & $0.497 \pm 0.057$ & $0.498 \pm 0.135$ & -- \\
      Female & Gene Score + Age + BMI + Smoke & $0.542 \pm 0.052$ & $0.441 \pm 0.138$ & -- \\
      Male & Age + BMI + Smoke & -- & -- & $0.656 \pm 0.083$ \\
      Male & Gene Score & -- & -- & $0.454 \pm 0.102$ \\
      Male & Gene Score + Age & -- & -- & $0.696 \pm 0.078$ \\
      Male & Gene Score + BMI & -- & -- & $0.481 \pm 0.100$ \\
      Male & Gene Score + Smoke & -- & -- & $0.414 \pm 0.102$ \\
      Male & Gene Score + Age + BMI & -- & -- & $0.671 \pm 0.090$ \\
      Male & Gene Score + Age + Smoke & -- & -- & $0.647 \pm 0.095$ \\
      Male & Gene Score + BMI + Smoke & -- & -- & $0.406 \pm 0.090$ \\
      Male & Gene Score + Age + BMI + Smoke & -- & -- & $0.634 \pm 0.091$ \\

\bottomrule
\end{tabularx}
\end{threeparttable}
\end{table}

\subsection*{AUROC in CKB Participants With CVD}

\begin{table}[H]
\centering
\caption{AUROC in CKB Participants With CVD}
\label{tab:ckb_auc_with_cvd}
\setlength{\tabcolsep}{3pt}
\renewcommand{\arraystretch}{1.05}
\footnotesize

\begin{threeparttable}
\begin{tabularx}{\textwidth}{
@{}
>{\centering\arraybackslash}p{0.08\textwidth}
>{\centering\arraybackslash}p{0.38\textwidth}
>{\centering\arraybackslash}X
>{\centering\arraybackslash}X
>{\centering\arraybackslash}X
@{}}
\toprule
\textbf{Group} &
\textbf{XGBoost Input} &
\thead{Breast Cancer\\AUROC (95\% CI)} &
\thead{Ovarian Cancer\\AUROC (95\% CI)} &
\thead{Prostate Cancer\\AUROC (95\% CI)} \\
\midrule
      All & Gene Score + Sex + Age + BMI + Smoke & $0.490 \pm 0.176$ & -- & $0.631 \pm 0.125$ \\
      Female & Age + BMI + Smoke & $0.469 \pm 0.179$ & -- & -- \\
      Female & Gene Score & $0.380 \pm 0.145$ & -- & -- \\
      Female & Gene Score + Age & $0.426 \pm 0.187$ & -- & -- \\
      Female & Gene Score + BMI & $0.437 \pm 0.177$ & -- & -- \\
      Female & Gene Score + Smoke & $0.379 \pm 0.158$ & -- & -- \\
      Female & Gene Score + Age + BMI & $0.440 \pm 0.189$ & -- & -- \\
      Female & Gene Score + Age + Smoke & $0.428 \pm 0.189$ & -- & -- \\
      Female & Gene Score + BMI + Smoke & $0.438 \pm 0.195$ & -- & -- \\
      Female & Gene Score + Age + BMI + Smoke & $0.455 \pm 0.181$ & -- & -- \\
      Male & Age + BMI + Smoke & -- & -- & $0.446 \pm 0.263$ \\
      Male & Gene Score & -- & -- & $0.663 \pm 0.267$ \\
      Male & Gene Score + Age & -- & -- & $0.805 \pm 0.113$ \\
      Male & Gene Score + BMI & -- & -- & $0.589 \pm 0.348$ \\
      Male & Gene Score + Smoke & -- & -- & $0.629 \pm 0.229$ \\
      Male & Gene Score + Age + BMI & -- & -- & $0.693 \pm 0.207$ \\
      Male & Gene Score + Age + Smoke & -- & -- & $0.774 \pm 0.084$ \\
      Male & Gene Score + BMI + Smoke & -- & -- & $0.529 \pm 0.243$ \\
      Male & Gene Score + Age + BMI + Smoke & -- & -- & $0.631 \pm 0.186$ \\

\bottomrule
\end{tabularx}
\end{threeparttable}
\end{table}

\section*{Supplementary ECG and Gene Preprocessing Details}

\begin{table}[htbp]
\centering
\caption{Summary of ECG-positive samples by gene}
\label{tab:ecg_positive_samples}
\begin{tabular}{lccc}
\hline
\textbf{Gene} & \textbf{Positive N} & \textbf{Positive ECG} & \textbf{Total (Female:Male)} \\
\hline
ATM   & 296 & 346 & 155:141 (52.4\% F) \\
BRIP1 & 153 & 172 & 76:77 (49.7\% F) \\
CHEK2 & 165 & 196 & 93:72 (56.4\% F) \\
PALB2 & 124 & 146 & 53:71 (42.7\% F) \\
BRCA1 & 71  & 80  & 21:50 (29.6\% F) \\
BRCA2 & 226 & 266 & 93:133 (41.2\% F) \\
\hline
\end{tabular}
\end{table}

\begin{table}[htbp]
\centering
\caption{Associations between Pathogenic Variants in Established Breast Cancer–Predisposition Genes and Risk of Breast Cancer in the UK Biobank}
\label{tab:gene_lp_p_distribution}
\resizebox{\textwidth}{!}{
\begin{tabular}{|c|c|c|c|c|}
\hline
\textbf{Gene LP/P} & 
\textbf{Case Patients $N = 2750$ (\%)} & 
\textbf{Controls $N = 43790$ (\%)} & 
\textbf{Odds Ratio (95\% CI)} & 
\textbf{$P$ value} \\
\hline
BRCA1 & 19 (0.6) \quad \textcolor{blue}{\textbf{0.85}} & 26 (0.06) \quad \textcolor{blue}{\textbf{0.11}} & 11.71 (6.47 -- 21.18) & 2.36E-12 \\
\hline
BRCA2 & 38 (1.3) \quad \textcolor{blue}{\textbf{1.29}} & 109 (0.2) \quad \textcolor{blue}{\textbf{0.24}} & 5.61 (3.87 -- 8.14) & 6.39E-15 \\
\hline
ATM & 26 (0.9) \quad \textcolor{blue}{\textbf{0.78}} & 168 (0.4) \quad \textcolor{blue}{\textbf{0.41}} & 2.4783 (1.64 -- 3.75) & 9.08E-05 \\
\hline
CHEK2 & 17 (0.6) \quad \textcolor{blue}{\textbf{1.08}} & 108 (0.2) \quad \textcolor{blue}{\textbf{0.42}} & 2.5159 (1.51 -- 4.20) & 0.0016 \\
\hline
PTEN & 1 & 41 & NA & NA \\
\hline
TP53 & 3 & 5 & NA & NA \\
\hline
6 All & 66 (3.9) \quad \textcolor{blue}{\textbf{4.08}} & 320 (1.0) \quad \textcolor{blue}{\textbf{1.2}} & -- & -- \\
\hline
\end{tabular}
}
\begin{tablenotes}
\footnotesize
\item Note: Data are presented as $n$ (\%) unless otherwise indicated. Blue values denote population-based epidemiological reference proportions reported in the UK Biobank population-based analysis of genes previously implicated in breast cancer, which included 32,247 case patients and 32,544 controls \cite{hu_population_breast_cancer}. NA denotes not applicable because there were too few events ($<5$) to calculate a stable odds ratio.
\end{tablenotes}
\end{table}

\section*{Demographic}

\begin{table}[H]
\centering
\caption{Baseline characteristics of participants in the training, validation, test, and external validation cohorts}
\label{tab:baseline_characteristics}
\small
\setlength{\tabcolsep}{3pt}
\renewcommand{\arraystretch}{1.15}

\resizebox{\textwidth}{!}{%
\begin{tabular}{
@{}
l
*{6}{c}
@{}}
\toprule
\textbf{Characteristic} &
\textbf{UKB Train} &
\textbf{UKB Validation} &
\textbf{UKB Test} &
\textbf{MIMIC} &
\textbf{-} & 
\textbf{CKB} \\
\midrule

Number of participants
& 47478 & 15826 & 15825 & 155721 & -- & 19278 \\

Age at ECG, mean (SD)
& 66.48 (7.77)
& 66.37 (7.72)
& 66.36 (7.82)
& 56.8 (19.4)
& --
& 58.3 (9.6)\\

Male sex, n (\%)
& 22648 (47.70\%)
& 7547 (47.69\%)
& 7592 (47.97\%)
& 73099 (46.9\%)
& --
& 6930 (35.9\%)\\

BMI, mean (SD)
& 26.62 (4.61)
& 26.73 (4.59)
& 26.60 (4.55)
& 28.8 (7.1)
& --
& 24.3 (3.5)\\

Smoking status, n (\%)
& 19260 (40.57\%)
& 6477 (40.93\%)
& 6416 (40.54\%)
& 8567 (5.5\%)
& --
& 4236 (22.0\%)\\

Breast cancer events, n (\%)
& 173 (0.36\%)
& 57 (0.36\%)
& 58 (0.37\%)
& 1068 (0.7\%)
& --
& 98 (0.79\%) \\

Ovarian cancer events, n (\%)
& 20 (0.04\%)
& 6 (0.04\%)
& 6 (0.04\%)
& 303 (0.2\%)
& --
& 19 (0.52\%) \\

Prostate cancer events, n (\%)
& 291 (0.61\%)
& 97 (0.61\%)
& 97 (0.61\%)
& 1394 (0.9\%)
& --
& 36 (0.52\%) \\

Race/ethnicity, n (\%)
& & & & & & \\

\hspace{1em}White
& 52371 (95.89\%)
& 17359 (95.80\%)
& 17508 (95.88\%)
& 83717 (53.76\%)
& --
& -- \\

\hspace{1em}East Asian
& 186 (0.34\%)
& 54 (0.30\%)
& 67 (0.37\%)
& 1711 (1.10\%)
& --
& 19278 (100\%)\\

\hspace{1em}Mixed
& 334 (0.61\%)
& 112 (0.62\%)
& 126 (0.69\%)
& 149 (0.10\%)
& --
& --\\

\hspace{1em}Asian or Asian British
& 764 (1.40\%)
& 251 (1.39\%)
& 239 (1.31\%)
& 2560 (1.64\%)
& --
& --\\

\hspace{1em}Black or Black British
& 475 (0.87\%)
& 171 (0.94\%)
& 144 (0.79\%)
& 16841 (10.81\%)
& --
& --\\

\hspace{1em}Others
& 481 (0.88\%)
& 168 (0.93\%)
& 173 (0.95\%)
& 11267 (7.24\%)
& --
& --\\

\hspace{1em}Missing
& 6 (0.01\%)
& 5 (0.03\%)
& 3 (0.02\%)
& 39476 (25.35\%)
& --
& --\\

\bottomrule
\end{tabular}%
}
\end{table}

\section*{ECG-Wide Association Analysis}

\begin{figure}[!ht] 
\centerline{\includegraphics[width=1\textwidth]{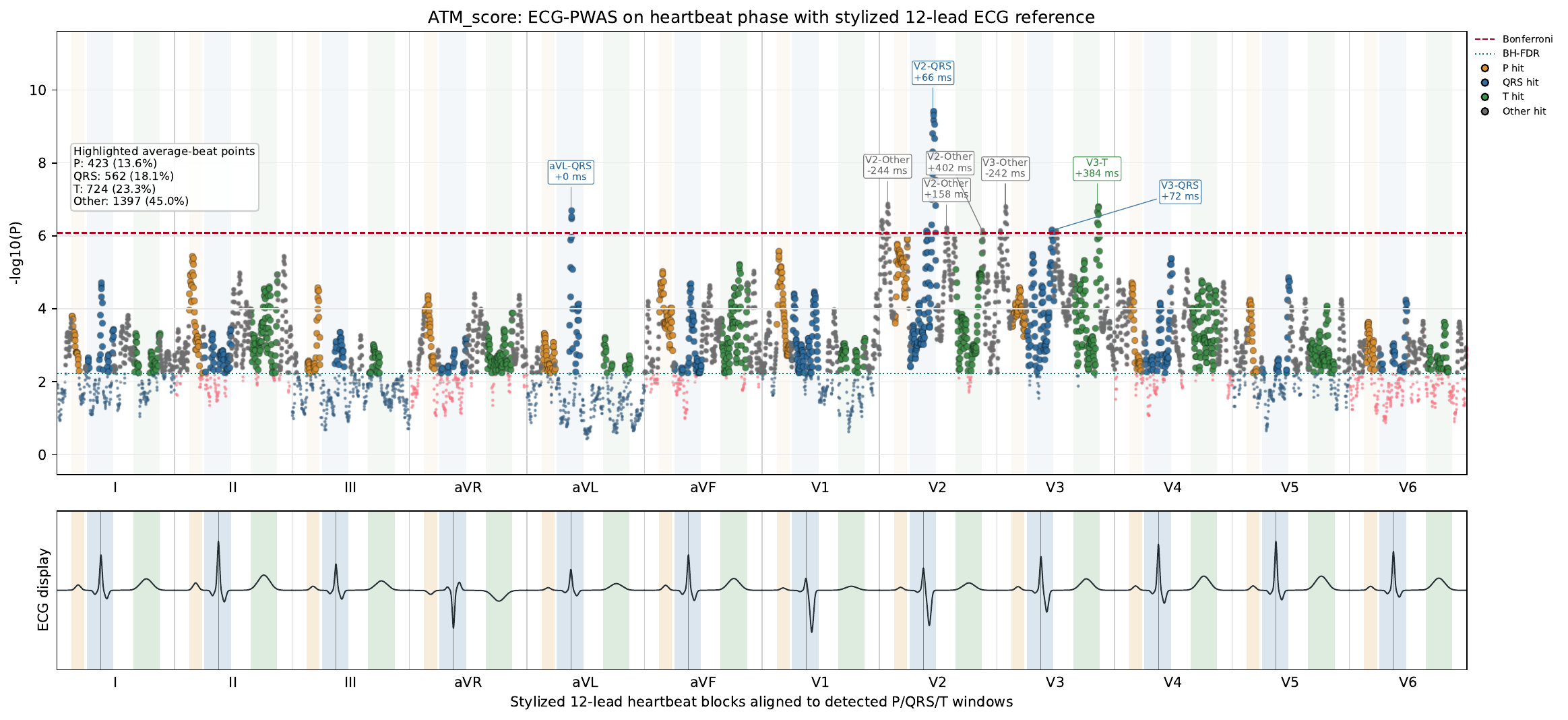}}
\caption{ATM gene score-associated ECG time points across 12 leads.
}
\label{pwas_ATM}
\end{figure}

\begin{figure}[!ht]
\centerline{\includegraphics[width=1\textwidth]{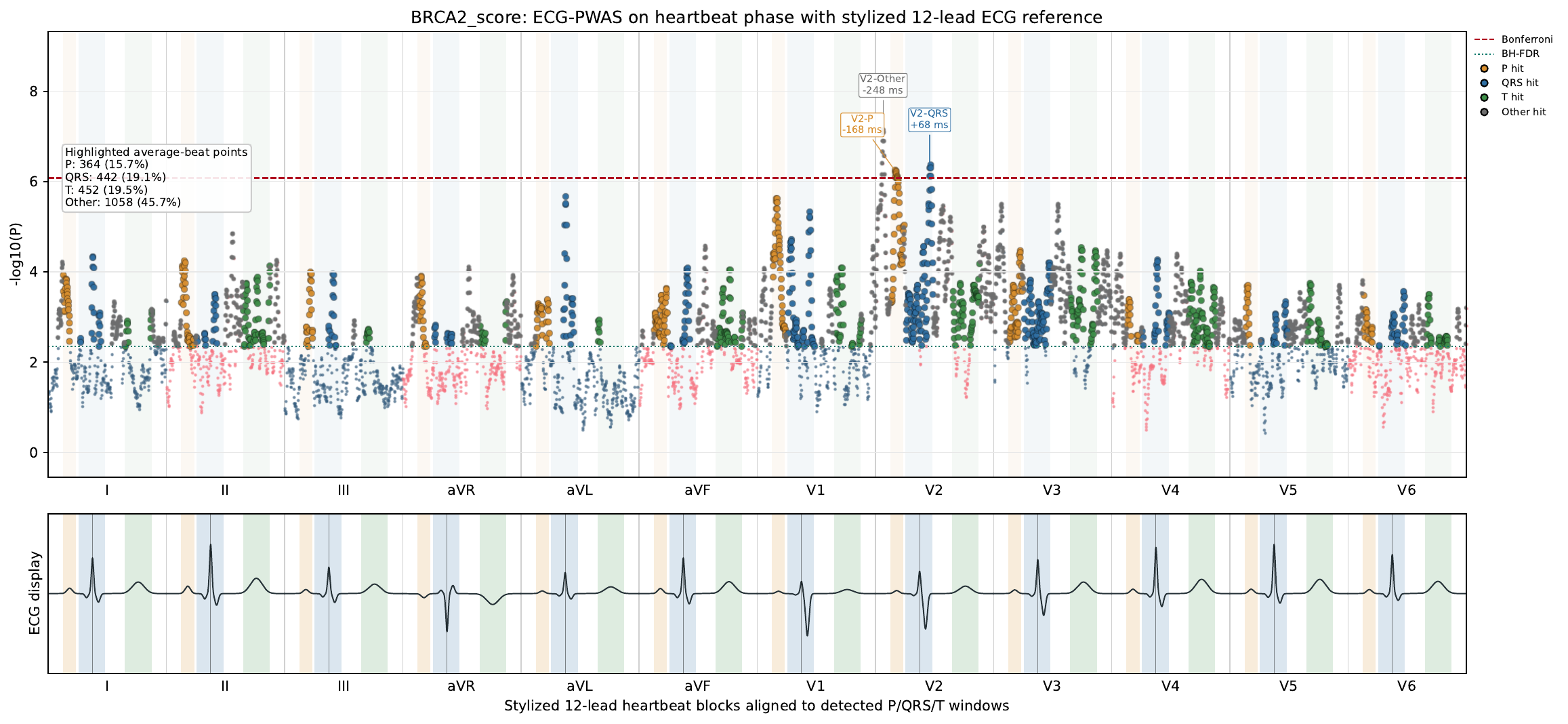}}
\caption{BRCA2 gene score-associated ECG time points across 12 leads.}
\label{pwas_BRCA2}
\end{figure}

\begin{figure}[!ht]
\centerline{\includegraphics[width=1\textwidth]{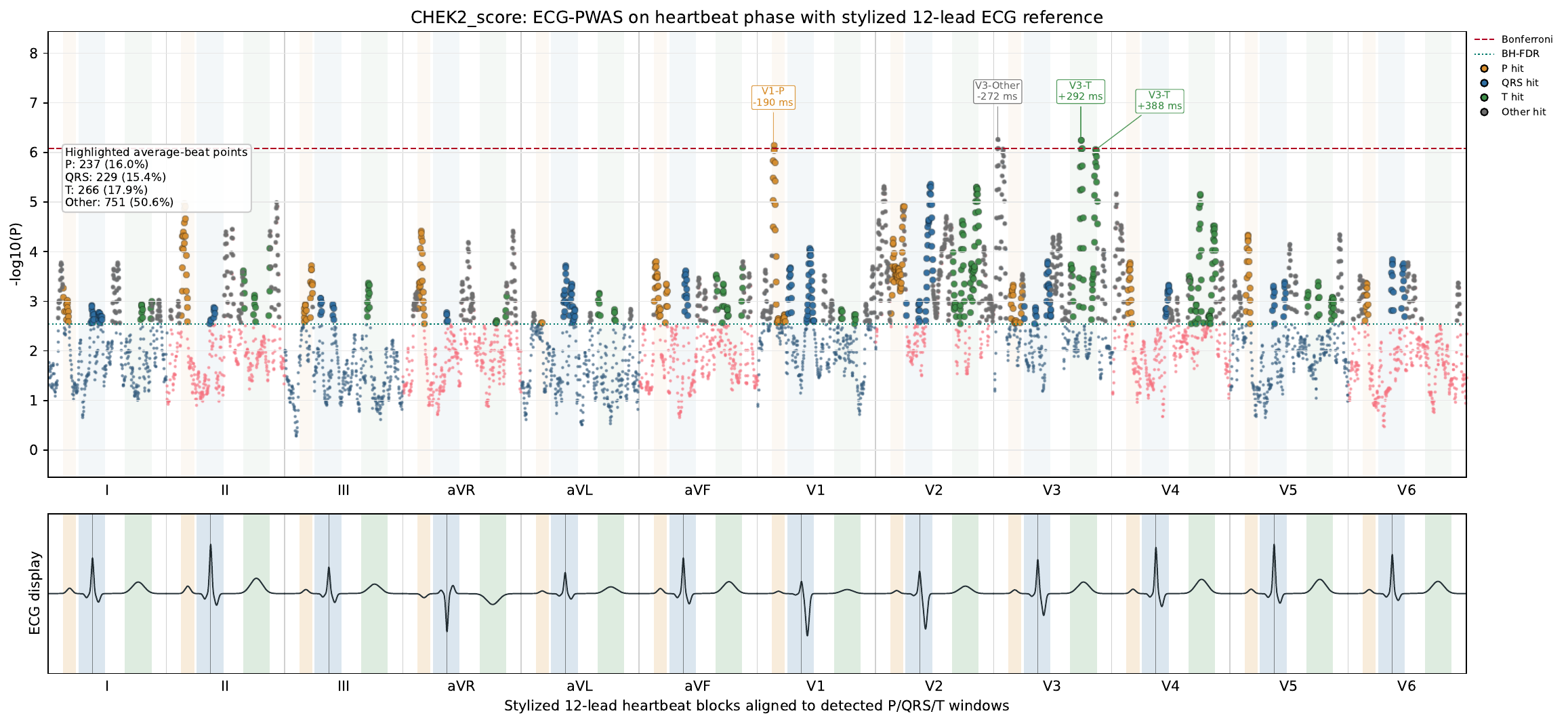}}
\caption{CHEK2 gene score-associated ECG time points across 12 leads.}
\label{pwas_CHEK2}
\end{figure}

\begin{figure}[!ht]
\centerline{\includegraphics[width=1\textwidth]{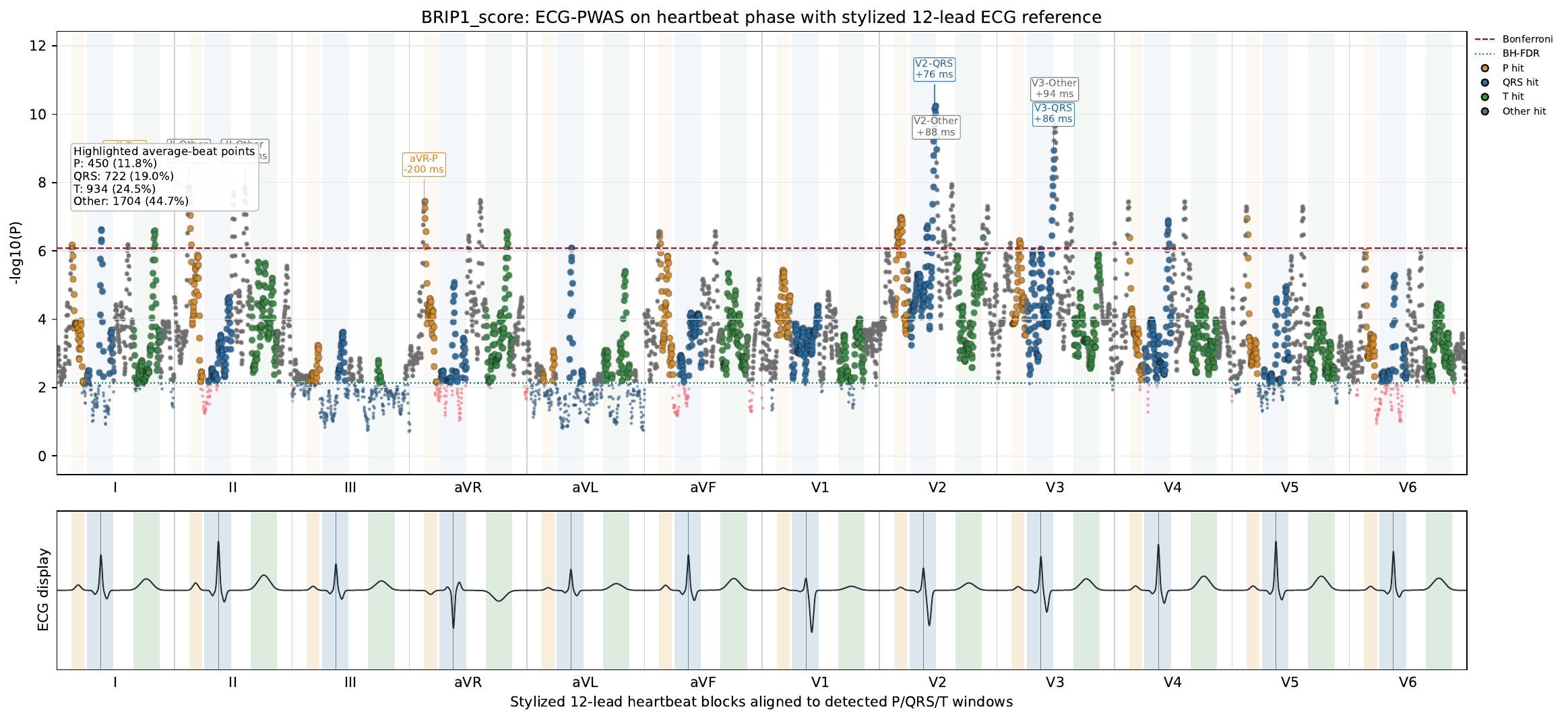}}
\caption{BRIP1 gene score-associated ECG time points across 12 leads.}
\label{pwas_BRIP1}
\end{figure}

\begin{figure}[!ht]
\centerline{\includegraphics[width=1\textwidth]{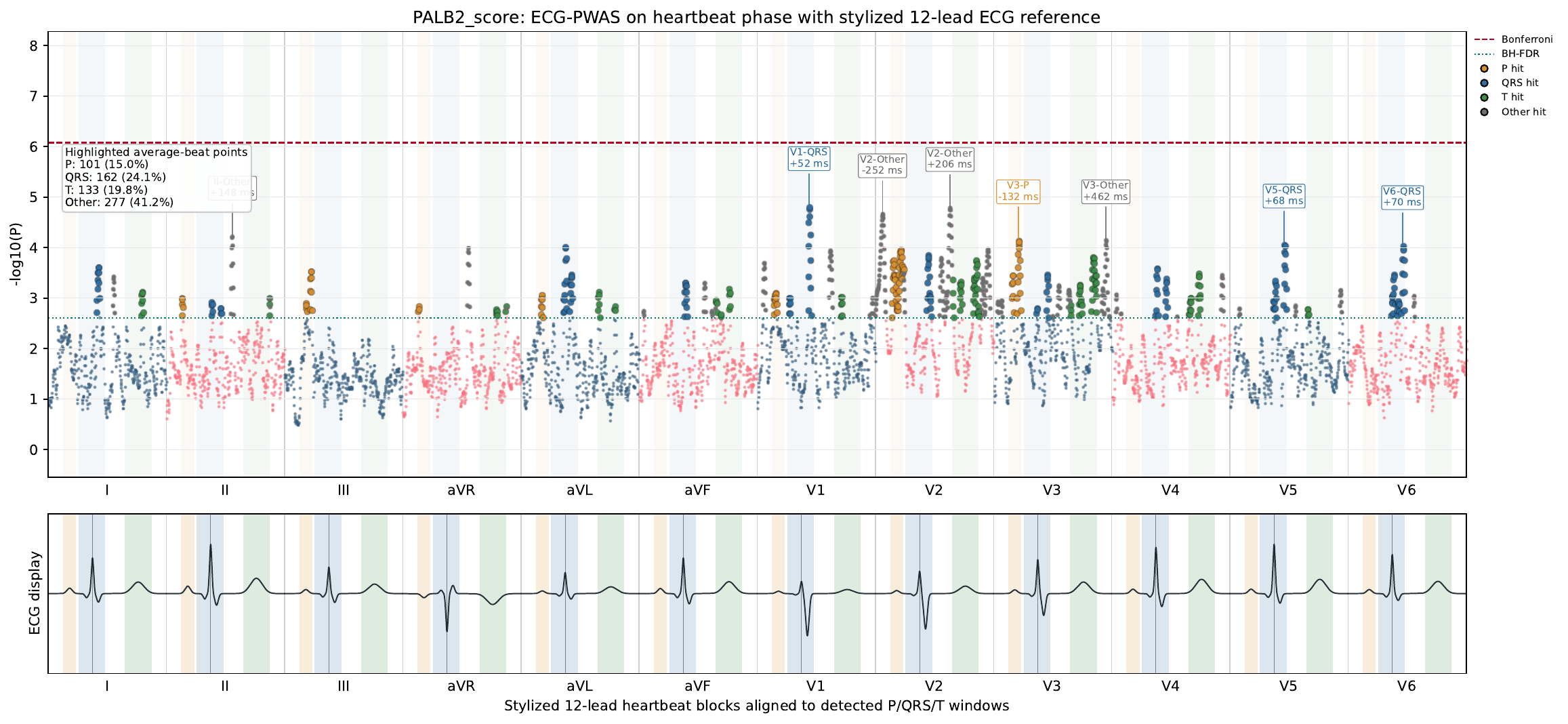}}
\caption{PALB2 gene score-associated ECG time points across 12 leads.}
\label{pwas_PALB2}
\end{figure}

\end{document}